\documentclass[UTF8,lettersize,journal]{IEEEtran}
\usepackage{amsmath,amsfonts,amssymb}
\usepackage{algorithmic}
\usepackage{algorithm}
\usepackage{array}
\usepackage[caption=false,font=normalsize,labelfont=sf,textfont=sf]{subfig}
\usepackage{textcomp}
\usepackage{stfloats}
\usepackage{url}
\usepackage{verbatim}
\usepackage{graphicx}
\usepackage{cite}
\usepackage{multirow}
\usepackage{booktabs}
\usepackage[justification=centering]{caption}
\usepackage{hyperref}
\usepackage{color}

\begin{document}

\title{Content-Induced Spatial-Spectral Aggregation Network for Change Detection in Remote Sensing Images}

\author{Yunlong Liu$^{*}$, Zekai Zhang
	\thanks{$^{*}$ Corresponding author. Y. Liu and Z. Zhang are with the School of Control Science and Engineering, Shandong University, Ji'nan 250061, China (e-mail: xxlnova@163.com, 202420810@mail.sdu.edu.cn).
}}

\markboth{Journal of \LaTeX\ Class Files,~Vol.~14, No.~8, August~2021}%
{Shell \MakeLowercase{\textit{et al.}}: A Sample Article Using IEEEtran.cls for IEEE Journals}


\maketitle

\begin{abstract}
The integration of spatial and spectral information is beneficial to the improvement of change detection performance. However, existing methods cannot efficiently suppress the influences of spatial and spectral differences in unchanged areas. To address these issues, in this paper we propose a content-guided spatial-spectral integration network (CSI-Net) for the fusion of global spatial details and spectral difference information. Specifically, the proposed CSI-Net is composed of a spatial reasoning (SR) module, a spectral difference (SD) module, and a content-guided integration (CGI) module. In the SR module, the spatial information is learned by cascaded graph convolution blocks for global modeling. The SD module is responsible for the extraction of spectral features, by calculating the means and variances of features to reduce the impact of spectral differences in unchanged regions. In addition, in order to integrate the spatial-spectral features efficiently, we design a CGI module to further take advantage of their complementary information. In this module, high-level content information is introduced as a guide for a proper interaction. Due to the efficient spatial-spectral fusion, the proposed CSI-Net can learn the changed features better while achieving a suppression of spectral differences. \textcolor{black}{Experimental results on LEVIR-CD, WHU-CD, and CLCD datasets demonstrate that the proposed CSI-Net produces better performance compared to state-of-the-art methods, and is applicable to different scenarios}. 
\end{abstract}

\begin{IEEEkeywords}
Spatial reasoning, spectral difference attention, content-guided integration, change detection, remote sensing.
\end{IEEEkeywords}

\section{Introduction}
\IEEEPARstart{C}{hange} \textcolor{black}{detection (CD) is intended to infer changes of interest, such as buildings, cropland, etc., occurring over multiple time periods in the same geographic area from multi-temporal remote sensing (RS) images \cite{Alpher60}} . This topic has attracted intensive attention owing to the realistic demand for many tasks, such as land-use regulation \cite{Alpher01} and biomass prediction \cite{Alpher02}. \textcolor{black}{With the development of imaging sensors and RS image fusion technology, \cite{Alpher66} many satellites, such as QuickBird and WorldView-4 have provided massive high-resolution (HR) RS images, which further promote the applications of CD.}

In recent years, deep neural networks (DNNs) have been widely developed and applied to the CD field because of their powerful representation ability. As one kind of the most popular architectures, convolution neural networks (CNNs) have gained a great attention. Many CD methods based on CNN \cite{Alpher03, Alpher04, Alpher05}, have been proposed and achieved competitive performance. For example, Peng \emph{et al.} \cite{Alpher06} proposed a semi-supervised CNN to integrate labeled and unlabeled data for training. \textcolor{black}{Yang \emph{et al.} \cite{Alpher07} introduced a pyramid architecture into CNNs to extract multi-scale features in multi-temporal images more efficiently.} To locate the boundaries of changed objects, Bai \emph{et al.} \cite{Alpher08} proposed an edge-guided CNN, in which structure information of images was used to facilitate the prediction of change maps. Although CNNs produce accurate results, these methods cannot capture global information in images due to the intrinsic locality of CNNs.

To better learn global information in multi-temporal images, researchers have considered various ways, such as nonlocal networks, transformer, and graph convolutional neural networks (GCNs) to obtain better CD results. For example, Lei \emph{et al.} \cite{Alpher09} introduced nonlocal modules into Siamese CNNs to model the global features in images. Zhang \emph{et al.} \cite{Alpher10} proposed a transformer-based CD method where the global relationships are extracted by a cross-temporal difference attention. Zhang \emph{et al.} \cite{Alpher11} combined CNN and transformer to simultaneously consider the local and global properties in images. Besides, as an effective means to model global relationships, GCNs are also applied to estimate the changed areas in multi-temporal images. \textcolor{black}{Wu \emph{et al.} \cite{Alpher12} employed a multi-scale GCN to pass the information in labeled nodes to unlabeled ones.} Saha \emph{et al.} \cite{Alpher13} also constructed a semi-supervised GCN, which utilized the superpixel segmentation technique for graph construction. Tang \emph{et al.} \cite{Alpher14} proposed an unsupervised CD GCN, which incorporated metric learning to obtain reliable pseudo-labels. Moreover, Liang \emph{et al.} \cite{Alpher15} considered the over-smoothing problem in GCNs and constructed a multi-scale fusion network based on GCN for CD. These methods mentioned above mainly focused on learning global property in terms of spatial information, whereas the spectral information in multi-temporal images is often either ignored or not exploited sufficiently.

According to \cite{Alpher16}, it is far from enough to infer changed areas accurately by exploring solely spatial information. Spatial and spectral information have to be considered jointly to produce satisfactory CD results. \textcolor{black}{Some CD methods that utilize spatial-spectral information simultaneously have emerged\cite{Alpher59}. For example, Zhang \emph{et al.} \cite{Alpher62} introduced the convolutional attention module (CBAM) to CD, which fused spatial attention and channel attention to enhance the difference information, thus obtaining better CD results. Wang \emph{et al.} \cite{Alpher63} designed SS-Former to extract spectral and spatial sequence information using different transformer encoders, and proposed T-Former to obtain temporal information. Zhan \emph{et al.} \cite{Alpher64} used 1D and 2D CNNs to extract spectral and spatial features, and converted the local tensor into a spectral-spatial vector, which reduced the CD complexity. But, the above methods mainly extract local spatial information through the attention mechanism and cannot model global spatial relationships.} The features extracted by GCN contain rich spatial global information. However lakes the spectral difference information between multi-temporal images, which makes DNNs sensitive to spectral changes caused by season or weather. For multi-temporal images, different collecting times may lead to some shifts in terms of spectral information of the same objects. For example, there are some spectral signature differences for vegetation areas in different seasons. The appearance features of the same buildings are sensitive to the imaging angle and lighting condition. The pseudo-changes caused by these shifts cannot be modified efficiently. Therefore, spatial and spectral information in images should be leveraged equally.

To deal with the above-mentioned issues, we propose a content-guided spatial-spectral integration network (CSI-Net) to learn more discriminative features for CD. \textcolor{black}{Specifically, we utilize residual networks (ResNets) as encoders to extract the multi-scale features in multi-temporal images. Then, a spatial reasoning (SR) module is designed to analyze the global spatial information, which is composed of cascaded graph convolution (GC) blocks.} In addition, we establish a spectral difference (SD) module to mitigate the interference of pseudo-change caused by estimating the mean and variance from features. To fuse the extracted spatial and spectral features by SR and SD modules, a content-guided feature integration (CGI) module is defined, in which the content information from encoders are regarded as a guide and ehnhanced by channel and spatial attention. \textcolor{black}{Finally, the output of the CGI module is gradually combined with the multi-scale feature extracted by the corresponding encoders to infer the change map. Compared to other methods that also utilize spatial and spectral information, the proposed CSI-Net integrates the content information to model global spatial relationships. In addition, we fully utilize the prior information of image styles rather than simple attention operations on the features. Finally, we propose the CGI module that facilitates the fusion of spatial and spectral information from different domains through the guidance of high-level semantic information, rather than direct concatenation.} The main above contributions of this paper are summarized as follows.

1) We propose a CSI-Net to integrate the spatial and spectral features in multi-temporal efficiently, in which the global relationships and spectral differences are jointly employed for the prediction of the change map. 

2) We design a SD module to enhance the extracted spectral feature. In this module, the mean and variance of features are computed and used to infer the spectral difference attention.

3) We construct a CGI module to better integrate the content information from encoders and the spatial and spectral features from SR and SD modules. \textcolor{black}{In this module, we introduce high-dimensional features enriched with semantic information as content information, which is further improved by an attention mechanism.}

To demonstrate the effectiveness of CSI-Net, we perform extensive experiments on LEVIR-CD, WHU-CD, and CLCD datasets. The proposed CSI-Net produces better performance in terms of subjective and objective evaluations when compared to some state-of-the-art methods.\textcolor{black}{Moreover, the proposed CSI-Net is applicable to many different scenarios and is able to exclude the effects of irrelevant changes on CD accuracy.}

\section{Related Work}
\subsection{Traditional CD Methods}
In order to address the needs of various applications, many CD methods based on traditional methods have been proposed in the last decades. For instance, change vector analysis (CVA) \cite{Alpher20} and compressed change vector analysis (C2VA) \cite{Alpher21} are employed to resolve the CD task by calculating the Euclidean distance among pixels. Nielsen \emph{et al.} \cite{Alpher22,Alpher23} proposed multivariate alteration detection (MAD), which was based on the criterion of maximizing the projected eigenvalue and variance. Some transformation-based methods, such as principal component analysis (PCA) \cite{Alpher24} and independent component analysis (ICA) \cite{Alpher25}, were used to infer change information between multi-temporal remote sensing images.\textcolor{black}{Zhang \emph{et al.}  \cite{Alpher67} utilized a low-rank representation to distinguish between changed and unchanged pixels, and took PCA to train a dictionary of changed and unchanged pixels to improve the accuracy of CD.} The above approaches are easy to implement. However despite some exception \cite{Alpher25}, they are mainly use only spectral information to model the changed features, ignoring the effect of spatial information.

\textcolor{black}{With the rise of machine learning, many spatial or sparse representation methods have been proposed to perform CD in multi-temporal images\cite{Alpher58}.} For example, Gong \emph{et al.} \cite{Alpher26} applied fuzzy clustering to CD for the estimation of changed regions. Erturk \emph{et al.} \cite{Alpher27} considered dictionary learning to learn the change pattern in images. In addition, K-nearest neighbors (KNN) \cite{Alpher28}, support vector machine (SVM) \cite{Alpher29}, and decision trees \cite{Alpher30} were also used to classify changed and unchanged pixels in multi-temporal images. \textcolor{black}{To exploit all the various types of features, Moser \emph{et al.} \cite{Alpher31} applied markov random field (MRF) to extract spatial information and incorporated it with spectral features.} Although these methods have been explored extensively, the limited representation ability of hand-crafted features makes them difficult to deal with complex scenes.

\subsection{DNN-Based CD Methods}
Nowadays, DNN-based CD methods have become the mainstream framework due to their excellent ability to leverage various features of images. For example, Daudt \emph{et al.} \cite{Alpher05} designed three fully convolutional (FC) network architectures (FC-EF, FC-Siam-conc, and FC-Siam-diff) with skip connections for CD. Feng \emph{et al.} \cite{Alpher32} introduced multi-scale feature interaction into the FC network to enhance the difference information between multi-temporal features. Liu \emph{et al.} \cite{Alpher33} also constructed an unsupervised style transformation FC framework to reduce task-irrelevant interference. Liu \emph{et al.} \cite{Alpher34} replaced the vanilla convolution in FC networks with depthwise separable convolution, by which the number of parameters was significantly reduced. \textcolor{black}{Lei \emph{et al.} \cite{Alpher35} embedded multi-scale decoupled convolution to FC networks for CD to improve its flexibility.}

For the further extraction of temporal information in images, recurrent neural networks (RNNs) have designed to extract the change patterns more efficiently by considering multi-temporal images as sequence data. For example, Mou \emph{et al.} \cite{Alpher36} first extracted features from multi-temporal images using a CNN and then mined the temporal correlation between these features through RNN. Bai \emph{et al.} \cite{Alpher37} embedded an adaptive weighting mechanism into long short-term memory (LSTM) to enhance the difference information between images. Chen \emph{et al.} \cite{Alpher38} employed multiple stacked RNNs to extract the sequence information in features for CD. To improve the consistency of images in terms of statistical distribution, generative adversarial networks (GANs) \cite{Alpher39} have considered for the adaptation of images from different times. Liu \emph{et al.} \cite{Alpher40} designed a supervised adaptive framework based on GAN to mitigate the impact of pseudo-changes on CD accuracy. Liang \emph{et al.} \cite{Alpher41} proposed a GAN-based semi-supervised method that simultaneously used labeled and unlabeled samples.

Recently, researchers have shown more interest in attention mechanisms. \textcolor{black}{Shi \emph{et al.} \cite{Alpher42} employed a CBAM to enhance the multi-temporal features and make them more discriminative.} Ding \emph{et al.} \cite{Alpher43} computed the spatial coherence and temporal correlation between images through the self-attention mechanism and achieved semantic reasoning. Li \emph{et al.} \cite{Alpher44} combined dense connections with hybrid attention to generate discriminative features by exploiting contextual information. Han \emph{et al.} \cite{Alpher45} constructed a hierarchical attention network using column attention and row attention to alleviate the computational complexity of attention inference.

As typical self-attention approaches to model global dependency, transformer-based CD methods have produced satisfactory results. For example, Zhang \emph{et al.} \cite{Alpher46} established a CD network to fuse global features, which only consists of transformers. Li \emph{et al.} \cite{Alpher47} utilized CNN and transformer to jointly extract local and global features. Very recently, as an emerging generative model, the diffusion model is also considered for CD in multi-temporal images. Bandara \emph{et al.} \cite{Alpher48} pre-trained a denoising diffusion probabilistic model (DDPM) on unlabeled images and then trained a lightweight detection network by using the features from DDPM. Daudt \emph{et al.} \cite{Alpher49,Alpher50} adopted a guided anisotropy diffusion algorithm to derive the CD results by iterative refinement.

\section{Proposed Method}
In this section, we first describe the overall architecture of the proposed CSI-Net, followed by a detailed description of the spatial reasoning module, the spectral difference module, the content-guided integration aggregation module, and the loss function with the related training procedure. 

\begin{figure*}[ht]
	\centering
	\includegraphics[scale=0.35]{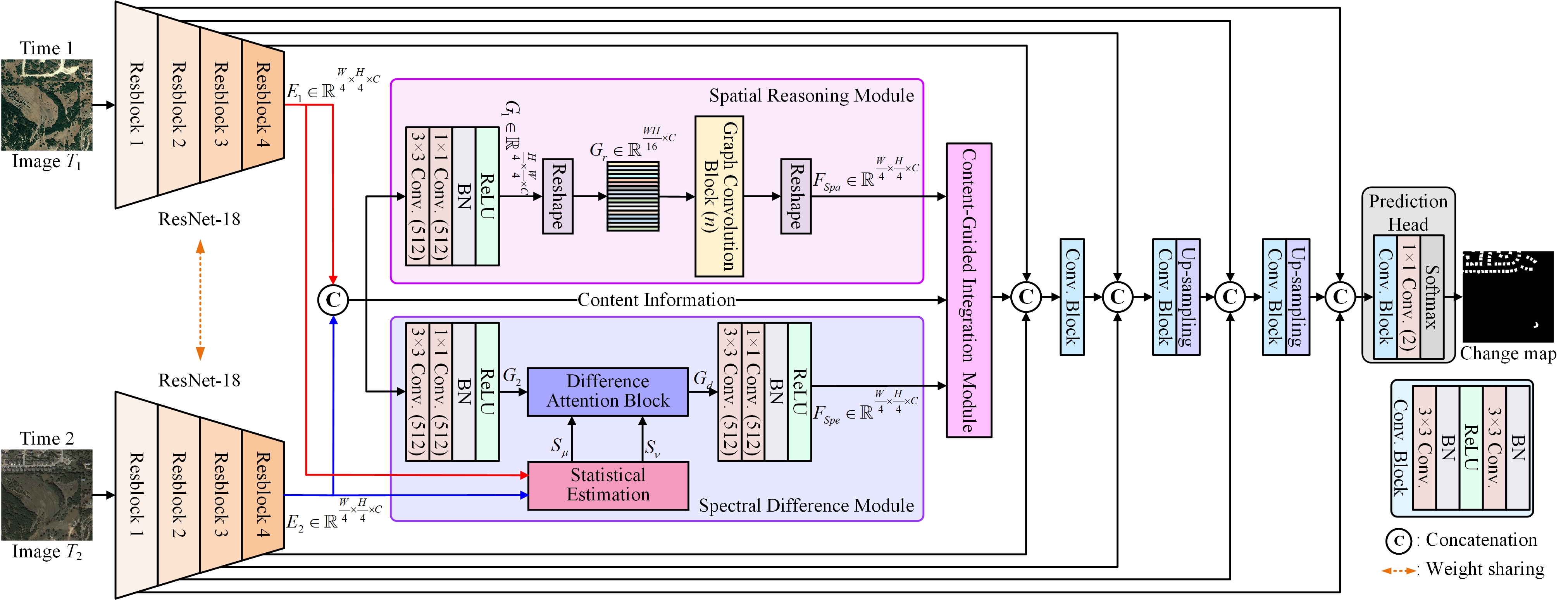}
	\captionsetup{font={footnotesize}}
	\caption{An architecture of proposed CSI-Net .}
	\label{fig1}
\end{figure*}

\subsection{Overall Architecture}

Fig. \ref{fig1} shows the overall framework of the proposed CSI-Net. \textcolor{black}{Multi-temporal images ${T_1} \in {\mathbb{R}^{{H} \times {W} \times 3}}$ and ${T_2} \in {\mathbb{R}^{{H} \times {W} \times 3}}$ are first projected by weight-shared Siamese encoders.} ${H}$ and ${W}$ represent the height and width of input images, respectively. In the Siamese encoder, a modified ResNet-18 \cite{Alpher51} is adopted. Specifically, we set the stride of the first ${7 \times 7}$ convolution layer in ResNet-18 as 1 and the following max-pooling layer is removed. In addition, we adjust the stride of all convolution layers in the last residual block to 1. \textcolor{black}{Through the settings, the four residual blocks in the modified ResNet-18 are used for the extraction of multi-scale features. Then, the outputs of the two encoders, ${E_1} \in {\mathbb{R}^{\frac{H}{4} \times \frac{W}{4} \times C}}$ and ${E_2} \in {\mathbb{R}^{\frac{H}{4} \times \frac{W}{4} \times C}}$, are fed into the SR module and SD module to learn the spatial relationships and spectral difference between multi-temporal images, respectively.} \emph{C} denotes the number of channels in the feature maps. Next, the extracted spatial and spectral information is fused by the CGI module, in which the content information of images is introduced to further enhance the change areas. \textcolor{black}{After that, the fused spatial and spectral features from the CGI module are progressively combined with multi-scale features from encoders by Conv. blocks and up-sampling layers with a bilinear operator.} Finally, the change map is produced by a prediction head consisting of a ${1 \times 1}$ convolution layer and Softmax.

\subsection{Spatial Reasoning Module}
In order to fully explore global spatial relationships, we design an SR module on the features of multi-temporal images. Specifically, the concatenation of ${E_1}$ and ${E_2}$ is condensed by a convolution block in the SR module for a primary fusion. The convolution block is composed of two convolution layers, a batch normalization (BN) layer and a rectified linear unit (ReLU). Then, the obtained ${G} \in {\mathbb{R}^{\frac{H}{4} \times \frac{W}{4} \times C}}$ is reshaped as ${G_r} \in {\mathbb{R}^{\frac{HW}{16}  \times C}}$ . To embed the similarity relationships into the features of multi-temporal images, ${G_r}$ is fed into ${n}$ stacked GC blocks. In the GC block, each row of ${G_r}$ is regarded as a node and their relationships are represented by an adjacency matrix. So, an adjacency matrix is first computed, which is defined as:
\begin{equation}
	\begin{aligned}
		\begin{array}{l}
			A = \phi \left( {{G_r}} \right) \cdot \phi {\left( {{G_r}} \right)^T}\\
			\phi \left( {{G_r}} \right) = {\rm{ReLU}}\left( {{G_r}{W_1} + {b_1}} \right)
		\end{array}
	\end{aligned}
\end{equation}
where ${\phi \left(  \cdot  \right)}$ is a linear layer with the weight ${W_1}$ and the bias ${b_1}$. Then, ${A} \in {\mathbb{R}^{\frac{HW}{16} \times \frac{HW}{16}}}$ is normalized by Softmax, and GC is achieved by combining the normalized adjacency matrix:
\begin{equation}
	\begin{aligned}
		{F_r}{\rm{ = ReLU}}\left( {\left( {\phi \left( {{G_r}} \right) + A\phi \left( {{G_r}} \right)} \right){W_2} + {b_2}} \right)
	\end{aligned}
\end{equation}
where ${F_r} \in {\mathbb{R}^{\frac{HW}{16} \times C}}$ is the result of GC containing the global information in images. ${W_2}$ and ${b_2}$ are learnable parameters in GC blocks. Finally, the output of the GC block ${F_{spa}} \in {\mathbb{R}^{\frac{H}{4} \times \frac{W}{4} \times C}}$ is generated by reshaping ${F_r}$. To properly learn the global relationships in multi-temporal images, ${n}$ GC blocks are cascaded for the enhancement of the spatial feature ${F_{spa}}$.

\subsection{Spectral Difference Module}
In the real CD task, due to the seasonal differences,and to the different obtain weather, and illumination conditions between different times, it is possible there exist some style shifts, whose influences should be removed from the extracted features\cite{1,2,3,4,5,6,7,8,9,10,11,12,13,zhang2026novel,zhang2023idd,zhang2025zero,zhang2024representation,zhang2026unification}. According to \cite{Alpher17,Alpher18,Alpher19}, the style of images is reflected by the mean and variance in feature space. For example, Bai \emph{et al.} \cite{Alpher18} calculated the mean and variance of features and achieved global style transfer by adaptive instance normalization (AdaIN).

Inspired by the modeling of image style, we consider the style shifts between multi-temporal images and design the SD module to alleviate this issue. Specifically, the proposed SD module consists of two convolution blocks, the statistical estimation of parameters of multi-temporal images block, and the DA block. ${E_1}$ and ${E_2}$ are fed into the statistical estimation block to obtain their style features, ${S_\mu } = \left\{ {S_\mu ^1,S_\mu ^2} \right\}$ and ${S_\nu } = \left\{ {S_\nu ^1,S_\nu ^2} \right\}$ as follows:
\begin{equation}
	\begin{aligned}
		S_\mu ^t\left( {i,j} \right) = \frac{1}{C}\sum\limits_{c = 1}^C {{E_1}\left( {i,j,c} \right)} 
	\end{aligned}
\end{equation}
\begin{equation}
	\begin{aligned}
		S_\nu ^t\left( {i,j} \right) = \frac{1}{C}\sum\limits_{c = 1}^C {{{\left( {{E_1}\left( {i,j,c} \right) - S_\mu ^1\left( {i,j} \right)} \right)}^2}}  
	\end{aligned}
\end{equation}
where ${S_\mu ^t} \in {\mathbb{R}^{\frac{H}{4} \times \frac{W}{4}}}$ and ${S_\nu ^t} \in {\mathbb{R}^{\frac{H}{4} \times \frac{W}{4}}}$ are the mean and variance of the image of time ${t}$ along channel direction, respectively.

\begin{figure}[ht]
	\vspace{-0.2cm}
	\centering
	\includegraphics[scale=0.83]{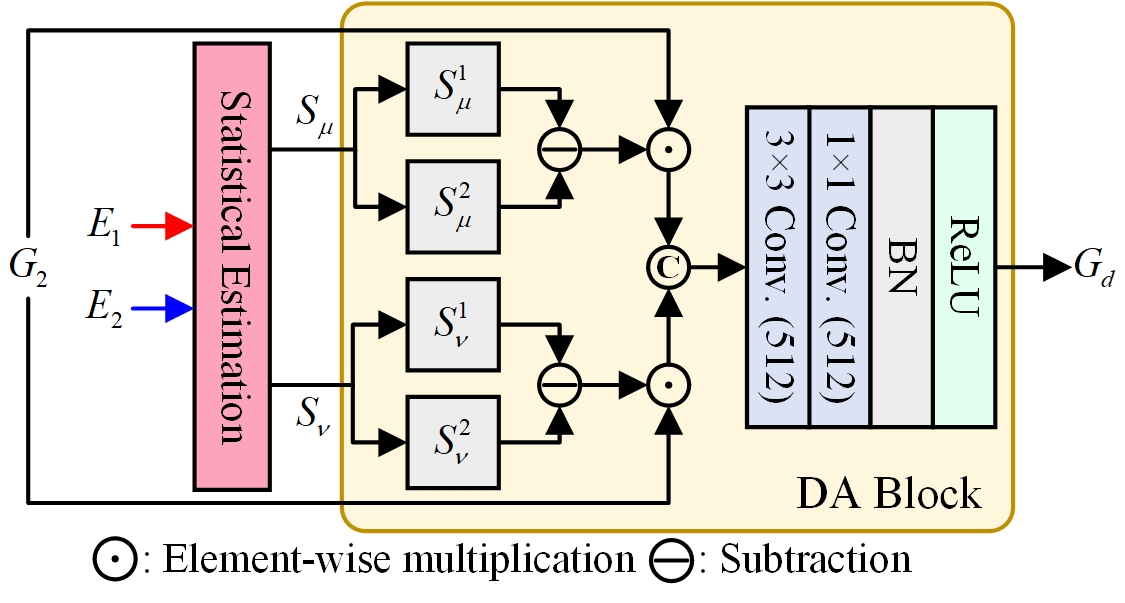}
	\captionsetup{font={footnotesize}}
	\caption{Architecture of the DA block.}
	\label{fig2}
	\vspace{-0.2cm}
\end{figure}

Then, the style features  ${S_\mu } = \left\{ {S_\mu ^1,S_\mu ^2} \right\}$ and ${S_\nu } = \left\{ {S_\nu ^1,S_\nu ^2} \right\}$ are combined with the features  ${E_1}$ and ${E_2}$  by the DA block shown in Fig. \ref{fig2}. In the SD module, the channels of concatenation of ${E_1}$ and ${E_2}$ are reduced to ${G_2} \in {\mathbb{R}^{\frac{H}{4} \times \frac{W}{4} \times C}}$ by the first convolution block. Furthermore, ${G_2}$ is further merged with the style differences of multi-temporal images in the DA block, which is formulated as:
\begin{equation}
	\begin{aligned}
		\begin{array}{l}
			{F_\mu } = {G_2} \odot \left( {S_\mu ^1 - S_\mu ^2} \right)\\
			{F_\nu } = {G_2} \odot \left( {S_\nu ^1 - S_\nu ^2} \right)
		\end{array} 
	\end{aligned}
\end{equation}
where ${ \odot }$ denote the element-wise multiplication. After concatenating ${F_\mu }$ and ${F_\nu }$, we set them as input to the following convolution block for the reduction of channels. By embedding the style differences into multi-temporal features, the attention in the DA block more efficiently highlights the features of changed areas and mitigates the impact of style shifts.

\subsection{Content-Guided Integration Module}
\begin{figure*}[ht]
	\vspace{-0.2cm}
	\centering
	\includegraphics[scale=0.8]{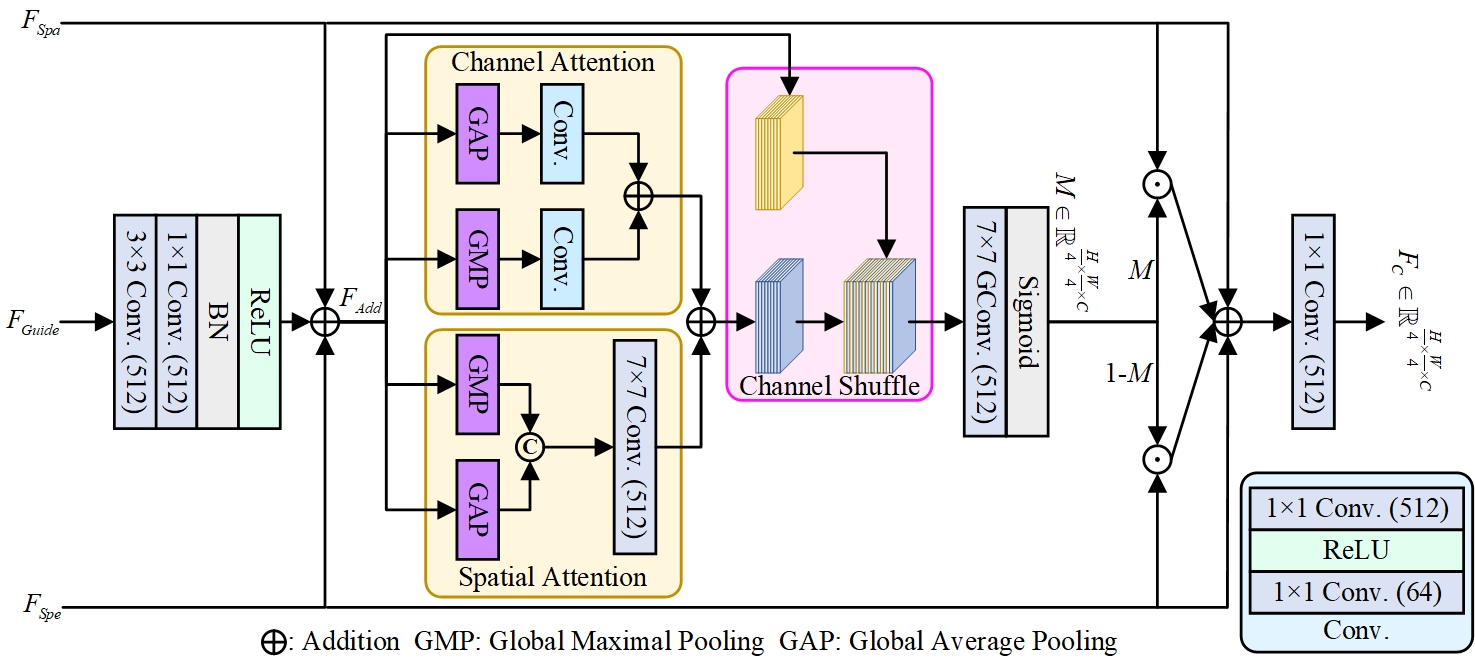}
	\captionsetup{font={footnotesize}}
	\caption{Architecture of the CGI module.}
	\label{fig3}
	\vspace{-0.2cm}
\end{figure*}

\textcolor{black}{According to\cite{Alpher61}, high-level features contain rich semantic information, which can better assist the localization of change regions and bridge the features of different domains. So, we take the high-level semantic features extracted from the backbone as the ``content`` to guide the fusion of spatial and spectral features.} In this section, we define the CGI module, in which the high-level features ${E_1}$ and ${E_2}$ from encoders are incorporated as the guide for a more efficient fusion between the spatial features from the SR module and the spectral features from the SD module. Fig. \ref{fig3} shows the architecture of the CGI module. In this module, the guide feature ${F_{Guide}}$ is projected by a convolution block and combined with ${F_{spa}}$ and ${F_{spe}}$. Then, the obtained feature ${F_{Add}}$ is enhanced by following channel and spatial attention blocks.

In the channel attention block, global average pooling (GAP) and global maximal pooling (GMP) along spatial dimensions are imposed on ${F_{Add}}$ to produce corresponding feature vectors with the length of \emph{C}. The two feature vectors from GAP and GMP are then expanded to the size of ${\frac{H}{4} \times \frac{W}{4} \times C}$ and projected by the following convolution blocks. Through the ${1 \times 1}$ convolution in these blocks, the relationships among channels are learned. In the spatial attention block, GAP and GMP are implemented on feature maps along channel dimension. The outputs of GAP and GMP have size ${\frac{H}{4} \times \frac{W}{4}}$. Then, they are concatenated and fed into a ${7 \times 7}$ convolution layer. In the spatial attention block, the details information in images is further enhanced for more efficient extraction of spatial features.

To integrate all attention maps, the outputs of channel and spatial blocks are added and combined with the content feature ${F_{Add}}$. To obtain a refined attention map, channel shuffle is used to extract the corresponding channels in different features and a ${7 \times 7}$ group convolution is applied to the two bundled channels. As shown in Fig. 3, there are \emph{C} groups and a channel-specific attention map is generated from each group by \emph{C} group convolution filters. Subsequently, to adaptively fuse ${F_{spa}}$ and ${F_{spe}}$, a weight map   is produced by a Sigmoid. Finally, the integrated spatial and spectral features are obtained as follows:
\begin{equation}
	\begin{aligned}
		{F_c} = conv\left( {{F_{spa}} + {F_{spe}} + {F_{spa}} \odot M + {F_{spe}} \odot \left( {1 - M} \right)} \right)
	\end{aligned}
\end{equation}
where ${conv\left(  \cdot  \right)}$ denotes the ${1 \times 1}$ convolution layer. Compared to general fusion operations, such as direct concatenation or addition, the content-guided fusion module improves the interaction of spatial and spectral information and avoids the mismatch in different feature spaces. So, the proposed CGI module fuses the spatial and spectral features more efficiently.

\subsection{Optimization}
To train the proposed CSI-Net, the binary cross-entropy (BCE) loss is used to calculate the difference between the predicted change map and the ground truth. It is define as:
\begin{equation}
	\begin{aligned}
		{L_{BCE}}\left( {\hat Y,Y} \right) =  - \sum {\left( {Y \odot \log \hat Y + \left( {1 - Y} \right)\log \left( {1 - \hat Y} \right)} \right)} 
	\end{aligned}
\end{equation}
where ${Y} \in {\mathbb{R}^{H \times W}}$ denote the ground truth and ${\hat Y} \in {\mathbb{R}^{H \times W}}$ is the predicted change map. When the value of the prediction map is 1, it corresponds to the changed pixel, and vice versa. The proposed CSI-Net is implemented through the PyTorch framework and trained on a server with an Intel Core i7-9700 processor, 3.0 GHz, and an NVIDIA GeForce RTX 2080ti (11GB). We use stochastic gradient descent (SGD) to optimize (7), and the initial learning rate is set to 0.001. Then, the learning rate decreases to 0 linearly until it is trained for 200 epochs. The batch size on all datasets is set to 4, and the training is terminated when the number of training epochs reaches 200.

\section{Experiments}
In this section, we first introduce the datasets for the experiments and the methods used for comparison. Then we show the results of comparisons and ablation experiments on three datasets to demonstrate the effectiveness of the proposed model. These datasets include LEVIR-CD \cite{Alpher53}, WHU-CD \cite{Alpher52}, and CLCD \cite{Alpher54}.

\subsection{Description of Datasets}
\emph{LEVIR-CD}: The dataset contains 637 pairs of RGB HR Google Earth image pairs with a resolution of 0.5m. \textcolor{black}{These bi-temporal images with a time span of 5 to 14 years show significant land-use changes, especially building growth, and decline.} Due to GPU memory limitation, we crop 637 image pairs with the size of $1024 \times 1024$ to non-overlapping $256 \times 256$ sub-image pairs. Then, we obtain a dataset including 7120/1024/2048 sub-image pairs for training/validation/testing, respectively.  

\emph{WHU-CD}: This dataset records the reconstruction of the Christchurch region of New Zealand since it experienced a 6.3 magnitude earthquake in 2011. \textcolor{black}{It consists of an aerial image pair with the size of $32507 \times 15345$ and a resolution of 0.2m. In this dataset, the main change of interest is buildings. R, G, and B bands are contained in the image pairs.} The dataset is partitioned into $256 \times 256$ sub-image pairs organized in 6069, 762, and 762 pairs for training, validation, and testing, respectively.

\emph{CLCD}: \textcolor{black}{The dataset is made up of 600 image pairs with a resolution of 0.5 to 2m, which contain changes of farmland, lakes, and buildings.} These images consist of three bands: red (R), green (G), and blue (B). The size is $256 \times 256$. We randomly divide these images into three parts:320/120/120 for training/validating/testing the proposed CSI-Net, respectively.

\emph{\textcolor{black}{Sensetime Dataset}}: \textcolor{black}{The dataset contains 2968 pairs of training images and 847 pairs of test images. We select 1000 image pairs from the training dataset and set up the training, validation, and test datasets in the ratio of 7:1:2. The image size in the dataset is 512×512 and the dataset contains six land cover categories, including water, ground, low vegetation, trees, buildings, and playgrounds, with a total of 31 “from-to” change types.} 

\subsection{Metrics}
In order to assess the CD performance of the proposed CSI-Net, we employ the following five standard evaluation metrics, which are precision (P), recall (R), F1-score (F1), intersection over union (IoU), and overall accuracy (OA). They are formulated as follows:
\begin{equation}
	\begin{aligned}
		\begin{array}{c}
			{\rm{R}} = {\rm{TP}} / {\rm{(TP+FN)}}\vspace{0.5ex} \\
			{\rm{P}} = {\rm{TP}} / {\rm{(TP+FP)}}\vspace{0.5ex} \\
			{\rm{F}}1 = {\rm{2PR}} /{\rm{(P+R)}}\vspace{0.5ex} \\
			{\rm{IoU}} = {\rm{TP}} / {\rm{(TP+FP+FN)}}\vspace{0.5ex} \\
			{\rm{OA}} = {\rm{(TP+TN)}} / {\rm{(TP+TN+FP+FN)}} \\
		\end{array}
	\end{aligned}
\end{equation}
where TP, TN, FP, and FN denote true positive, true negative, false positive, and false negative, respectively. For all metrics, larger represent better results.

\begin{figure*}[ht]
	\vspace{-0.2cm}
	\centering
	\includegraphics[scale=0.21]{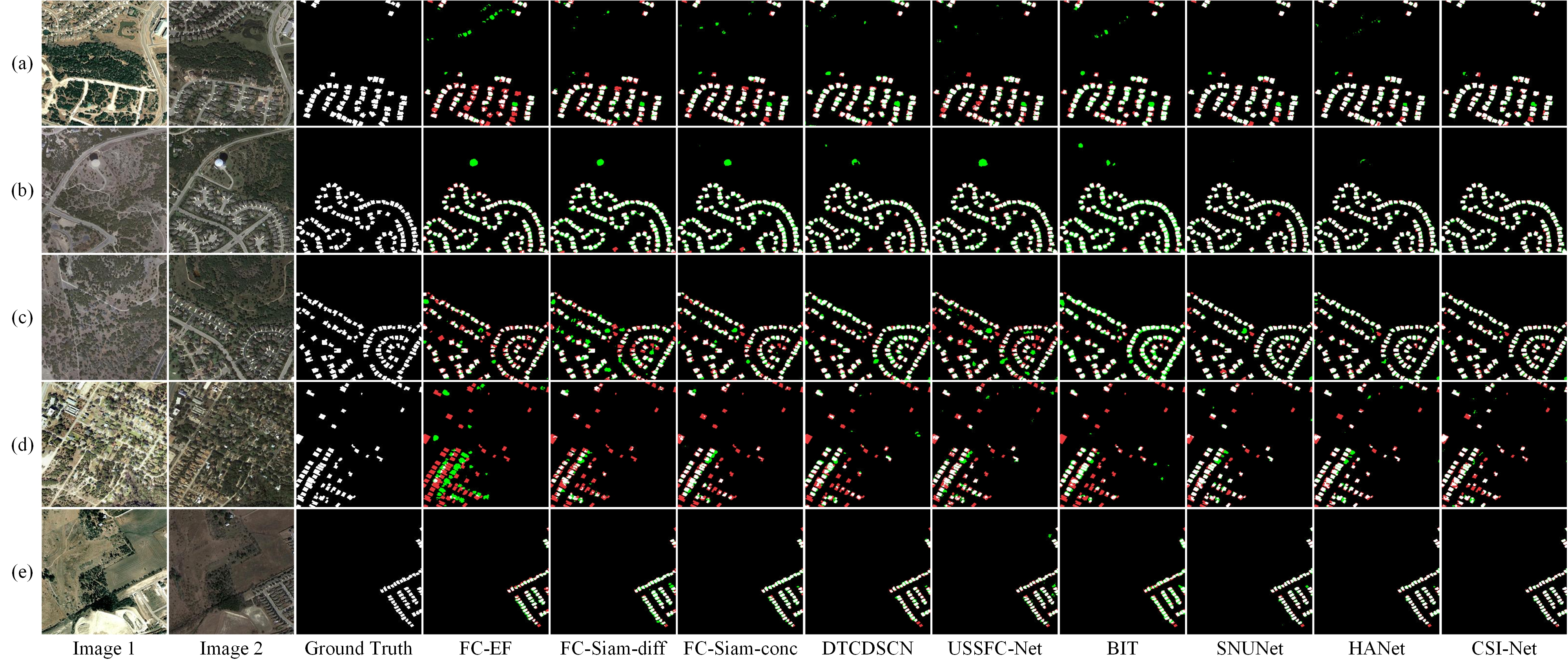}
	\captionsetup{font={footnotesize}}
	\caption{\textcolor{black}{Qualitative comparison of all methods on the LEVIR-CD dataset. (a)-(e): Prediction results of all methods on examples of different image pairs.}}
	\label{fig4}
	\vspace{-0.3cm}
\end{figure*}

\subsection{Compared Methods and Implementation Details}
To prove the effectiveness of the proposed CSI-Net,we choose eight state-of-the-art methods for comparison, which are the {FC-EF} \cite{Alpher05}, the {FC-Siam-conc} \cite{Alpher05}, the {FC-Siam-diff} \cite{Alpher05}, the dual task constrained deep Siamese convolutional network {(DTCDSCN)} \cite{Alpher55}, \textcolor{black}{the Ultralightweight Spatial--Spectral Feature Cooperation Network (USSFC-Net) \cite{Alpher35},} the bi-temporal image transformer (BIT) \cite{Alpher56}, the combination of Siamese network and NestedUNet (SNUNet) \cite{Alpher04}, and the hierarchical attention network (HANet) \cite{Alpher46}. All methods are briefly described bellow:

\emph{FC-EF}: A fully convolutional neural network with a U-Net adopts an early fusion approach to process bi-temporal images, which are concatenated and directly served as the input to the network.

\emph{FC-Siam-conc}: A Siamese fully convolutional neural network that applies a post-fusion strategy to extract features from bi-temporal images. In this method, they are integrated together as input of the decoder and multi-level features are introduced by a skip connection.

\emph{FC-Siam-diff}: A Siamese fully convolutional difference neural network is considered. The model first calculates the absolute value of the feature differences from the encoder. Then, a skip connection is linked with the corresponding layer of the decoder.

\emph{DTCDSCN}: A Siamese convolutional neural network based on dual-task constraints is constructed and its performance is improved by introducing dual-attention modules and modified loss functions.

\textcolor{black}{\emph{USSFC-Net}: The spatial-spectral attention weights are obtained without additional parameters by introducing an efficient spatial and spectral feature coordination strategy.}
	
\emph{BIT}: Semantic tokens are obtained by segmenting the bi-temporal features. The semantic context information is extracted by a Transformer encoder, in which the context-rich tokens are fed back to the feature space for the enhancement of CD results.

\emph{SNUNet}: \textcolor{black}{The information loss caused by downsampling is mitigated by a dense network, meanwhile the information interaction between multi-scale features is enhanced.} The ensemble channel attention module is used to constrain the information fusion.

\emph{HANet}: A hierarchical attention network incorporates parallel convolution and a simplified self-attention mechanism for feature integration and refinement.

To guarantee fairness, the number of epochs for all comparison methods is also set to 200 and all methods have the same batch size as CSI-Net.

\begin{figure*}[ht]
	\centering
	\includegraphics[scale=0.21]{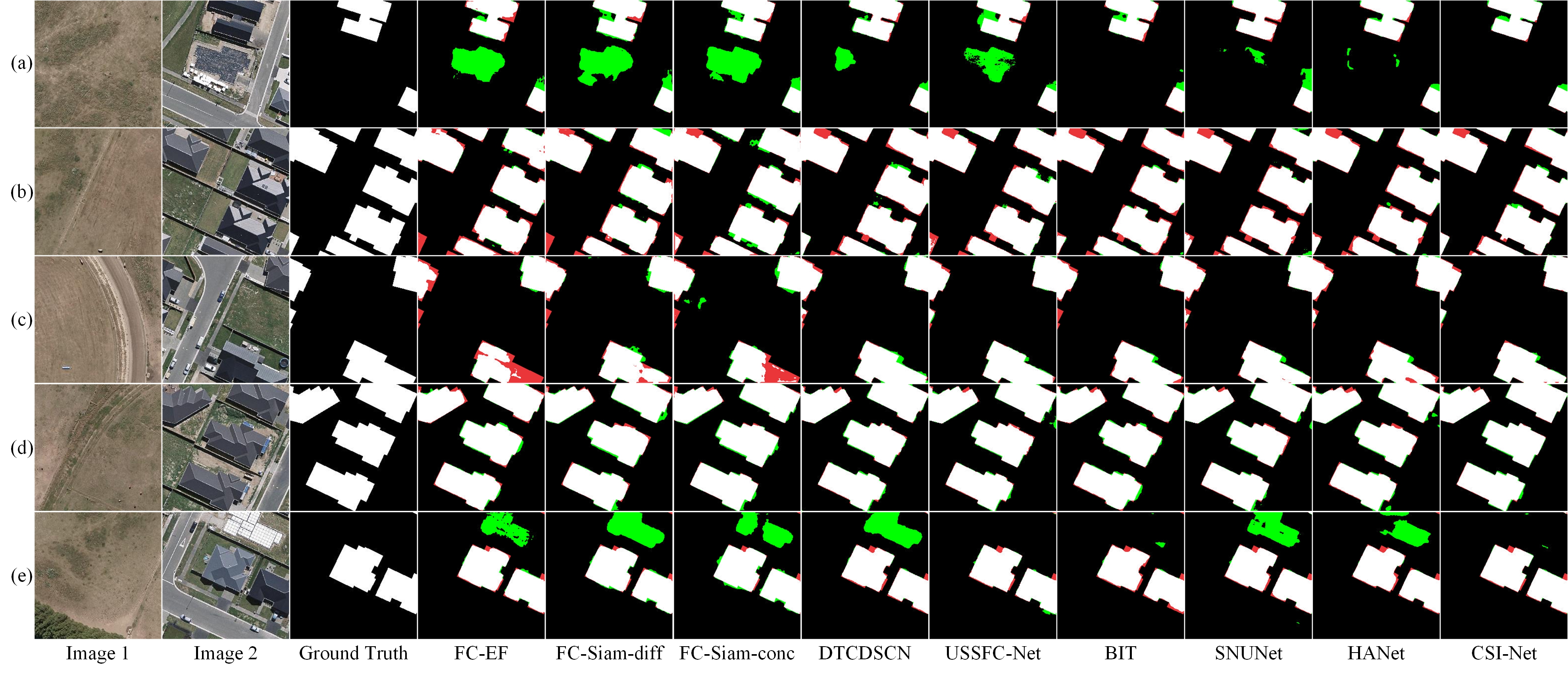}
	\captionsetup{font={footnotesize}}
	\caption{\textcolor{black}{Qualitative comparison of all methods on the WHU-CD dataset. (a)-(e): Prediction results of all methods on examples of different image pairs.}}
	\label{fig5}
	\vspace{-0.3cm}
\end{figure*}

\subsection{Results on the Comparison State-of-the-Art Methods}
\emph{1) Experiments on the LEVIR-CD Dataset}: Examples of CD results of all methods on the LEVIR-CD dataset are shown in Fig. \ref{fig4}, where, TP, TN, FP, and FN are presented in white, black, green, and red, respectively. \textcolor{black}{The LEVIR-CD dataset focuses on changes of buildings, and it records both additions and deletions of buildings. }As shown in Fig. \ref{fig4}(a), the change regions are intensively distributed. Many small change regions are not detected by the {FC-EF} and the {FC-Siam-diff}. (regions labeled in red) \textcolor{black}{{While many unchanged regions in the results of USSFC-Net, {BIT}, {SNUNet}, and {HANet} are improperly identified as changed.}} (marked in green) In Fig. \ref{fig4}(b), there is a pseudo-change region caused by the high squint of imaging system, which is mistakenly recognized as a change region by all compared methods except the proposed CSI-Net. In Figs. \ref{fig4} (c) and (e), there are some change regions consisting of multiple adjacent buildings. The detection results of {DTCDSCN} and {FC-Siam-conc}, tend to be connected without clear boundaries between buildings. From Fig. \ref{fig4}(d), we can see that the results of FC-EF, FC-Siam-diff, and DTCDSCN all degrade when the scenes are more complex, and these methods fail to describe the correct change regions. As shown in Fig. \ref{fig4}, all change areas in the examples are caused by new buildings, and the distribution of change areas is fragmented and imbalanced above them, an accurate global spatial relationship is achieved by the SR module in the proposed CSI-Net, which produces more accurate results.

\begin{table}[h]
	\centering
	\captionsetup{font={footnotesize}}
	\caption{\textsc{\textcolor{black}{Quantitative Results of All Methods on the LEVIR-CD Dataset.}}}
	\footnotesize
	\renewcommand\arraystretch{1.2}
	\begin{tabular}{cccccc}
		\hline
		Metric (\%)  & P              & R              & F1             & IoU            & OA             \\ \hline
		FC-EF        & 89.92          & 83.55          & 86.42          & 78.18          & 97.58          \\
		FC-Siam-conc & 88.40          & 86.85          & 87.61          & 79.72          & 97.65          \\
		FC-Siam-diff & 89.15          & 82.80          & 85.66          & 77.21          & 97.45          \\
		DTCDSCN      & 90.61          & 84.33          & 87.18          & 79.17          & 97.71         \\
		\textcolor{black}{USSFC-Net}      & \textcolor{black}{90.44}          & \textcolor{black}{86.07}          &  \textcolor{black}{88.12}          & \textcolor{black}{80.42}          & \textcolor{black}{97.83}         \\
		BIT          & 87.22          & 86.10          & 86.65          & 78.45          & 97.46          \\
		SNUNet       & \textbf{94.29} & 89.32          & 91.64          & 85.47          & 98.47          \\
		HANet & 93.89          & 89.92          & 91.80          & 85.71          & 98.49          \\
		Proposed CSI-Net   & 93.98          & \textbf{90.59} & \textbf{92.21} & \textbf{86.34} &  \textbf{98.55} \\ \hline
	\end{tabular}
    \label{tableI}
\vspace{-0.2cm}
\end{table}

\begin{figure*}[ht]
	\centering
	\includegraphics[scale=0.21]{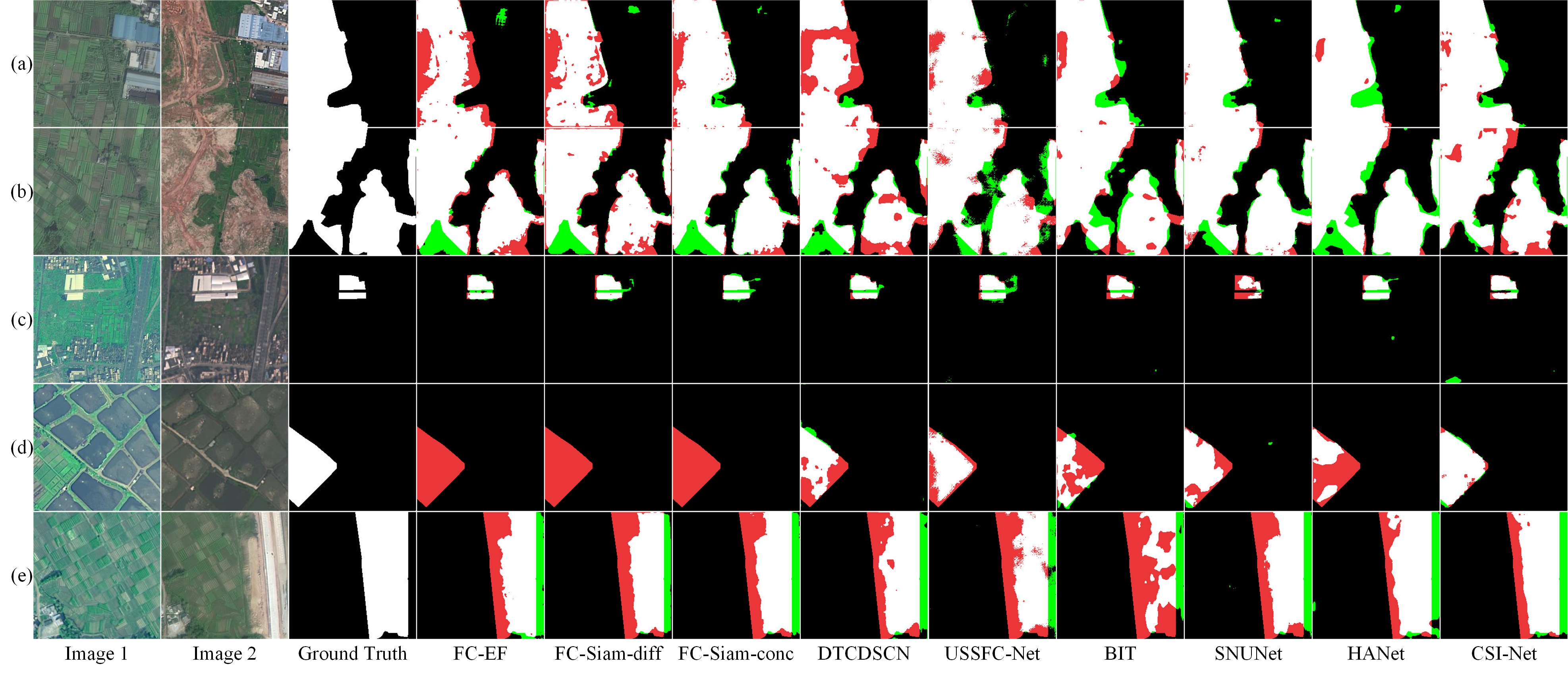}
	\captionsetup{font={footnotesize}}
	\caption{\textcolor{black}{Qualitative comparison of all methods on the CLCD dataset. (a)-(e): Prediction results of all methods on examples of different image pairs.}}
	\label{fig6}
	\vspace{-0.3cm}
\end{figure*}

Table \ref{tableI} reports the qualitative results of all methods on the LEVIR-CD dataset. The optimal values are highlighted in bold. We can see that the proposed CSI-Net achieves the best results in terms of four metrics, R, F1, IOU, and OA, while the optimal result for P is obtained by SNUNet. However, the P indicator of CSI-Net is only 0.31$\%$ smaller than SNUNet, whereas both IoU and F1 are sharply higher. \textcolor{black}{The overall accuracy of USSFC-Net, which also utilizes spatial spectral information, is much lower than the proposed CSI-Net}

\begin{table}[ht]
	\centering
	\captionsetup{font={footnotesize}}
	\caption{\textsc{\textcolor{black}{Quantitative Results of All Methods on the WHU-CD Dataset.}}}
	\footnotesize
	\renewcommand\arraystretch{1.2}
	\begin{tabular}{cccccc}
		\hline
		Metric (\%)  & P              & R              & F1             & IoU            & OA             \\ \hline
		FC-EF        & 93.20          & 90.21          & 91.64          & 85.50          & 98.64          \\
		FC-Siam-conc & 89.17          & 91.63          & 90.36          & 83.59          & 98.33          \\
		FC-Siam-diff & 79.71          & 92.22          & 87.60          & 79.71          & 97.66          \\
		DTCDSCN      & 94.96          & 88.75          & 91.60          & 85.41          & 98.48         \\
		\textcolor{black}{USSFC-Net}      & \textbf{\textcolor{black}{96.60}}          & \textcolor{black}{92.77}          & \textcolor{black}{94.59}          & \textcolor{black}{90.16}          & \textcolor{black}{99.13}         \\
		BIT          & 93.31          & 94.46          & 93.90          & 89.01          & 98.96          \\
		SNUNet       & 93.24          & 94.18          & 93.70          & 88.70          & 98.93          \\
		HANet & 95.21          & 94.55          & 94.88          & 90.63          & 99.14          \\
		Proposed CSI-Net   & 96.27         & \textbf{95.43} & \textbf{95.85} & \textbf{92.28} &  \textbf{99.31} \\ \hline
	\end{tabular}
    \label{tableII}
\vspace{-0.3cm}
\end{table}

\emph{2) Experiments on the WHU-CD Dataset}: Fig. \ref{fig5} shows examples of the results obtained by all methods on the WHU-CD dataset. \textcolor{black}{This dataset primarily records the growth of buildings, ignoring changes of roads, vegetation, and so on. It contains approximately 22,000 individual buildings. }In this dataset, the buildings are larger and have sharper edges than in the LEVIR-CD dataset due to the higher resolution.  \textcolor{black}{From Figs. \ref{fig5}(a) and (e), one can easily see that FC-EF, FC-Siam-diff, FC-Siam-conc, DTCDSCN, USSFC-Net, SNUNet, and HANet infer the unchanged regions as the changed ones, with large green areas of false positive.} As we can see in Fig. \ref{fig5}(b) and (d), the proposed CSI-Net produces maps with more accurate edges and extracts more complete change regions compared to BIT and other methods. In greater detail in Fig. \ref{fig5}(c), one can observe that FC-EF, FC-Siam-diff, and FC-Siam-conc all judge the part of the building in the bottom-right corner as an unchanged area, which leads to fragmentary CD results.

Table \ref{tableII} illustrates the metric values obtained by all the considered methods. \textcolor{black}{USSFC-Net achieved the highest P. The proposed CSI-Net obtains the best R, F1, IoU, and OA.} Thus, the comprehensive performance of the proposed CSI-Net is better than those of the comparison methods.\textcolor{black}{ Most of the metrics considered are higher than the values obtained by other methods.} Specifically, IoU is improved by 1.65$\%$ over the second best of HANet, and CSI-Net also produces higher values than HANet for the other four metrics.

\emph{3) Experiments on the CLCD Dataset}: \textcolor{black}{ Fig. \ref {fig6} shows the results of all methods on some examples of the CLCD dataset, which contains images with a more complex background and a lower resolution, and its main record of changes in arable land also contains changes in lakes and buildings. }In addition, this dataset mainly contains changes in terms of land use, and change areas are more similar to the background compared to those in the LEVIR-CD and the WHU-CD datasets. This scenario manufactures the difficulty to detect CD areas accurately for all methods. For example, in Figs. \ref {fig6}(a)-(d), many large-scale change regions are not detected by FC-EF, FC-Siam-conc, and FC-Siam-diff, this may be caused by the limited spectral differences in these images. Compared to other methods, the proposed CSI-Net has better detection results because the spectral difference information is fully exploited use of by the SD module.

Table \ref{tableIII} lists the quantitative results obtained by all methods. The proposed CSI-Net performs best in terms of all metrics. This confirms it has better overall performance.

\begin{figure*}[ht]
	\centering
	\includegraphics[scale=0.315]{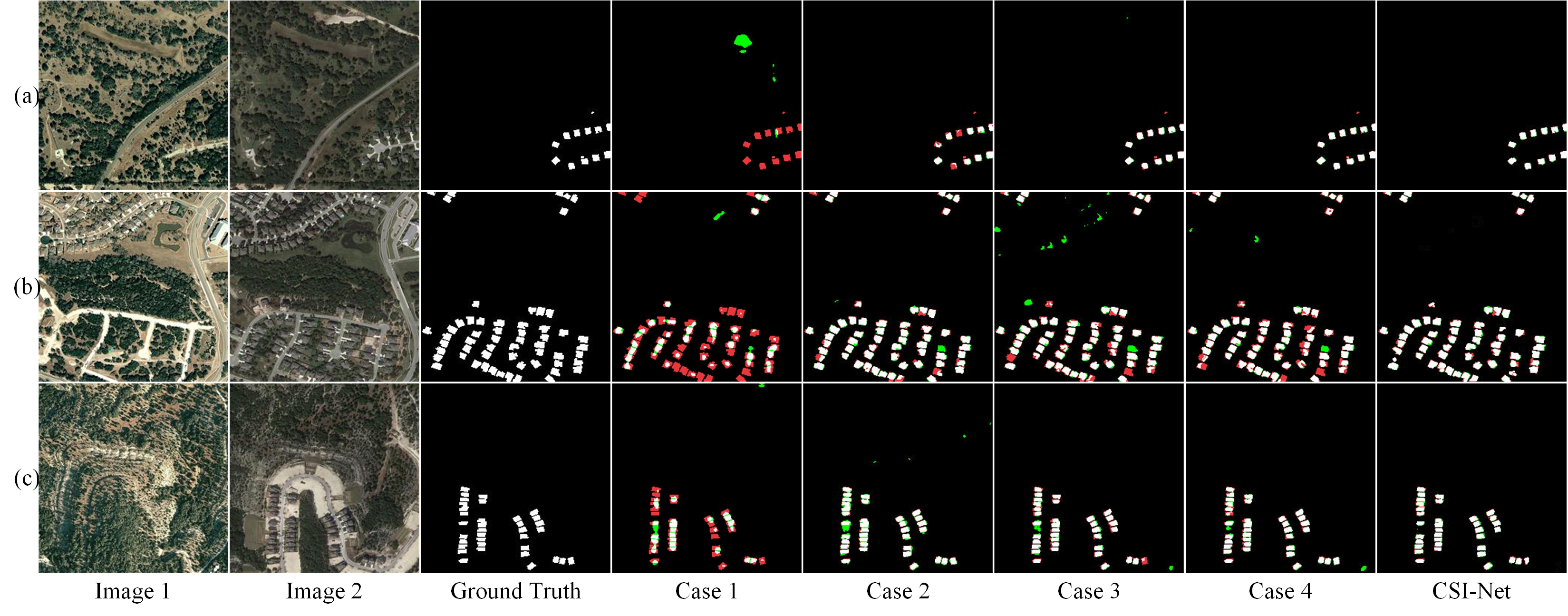}
	\captionsetup{font={footnotesize}}
	\caption{Examples of qualitative results of the ablation study on the LEVIR-CD dataset.}
	\label{fig7}
	\vspace{-0.2cm}
\end{figure*}

\begin{figure*}[ht]
	\centering
	\includegraphics[scale=0.315]{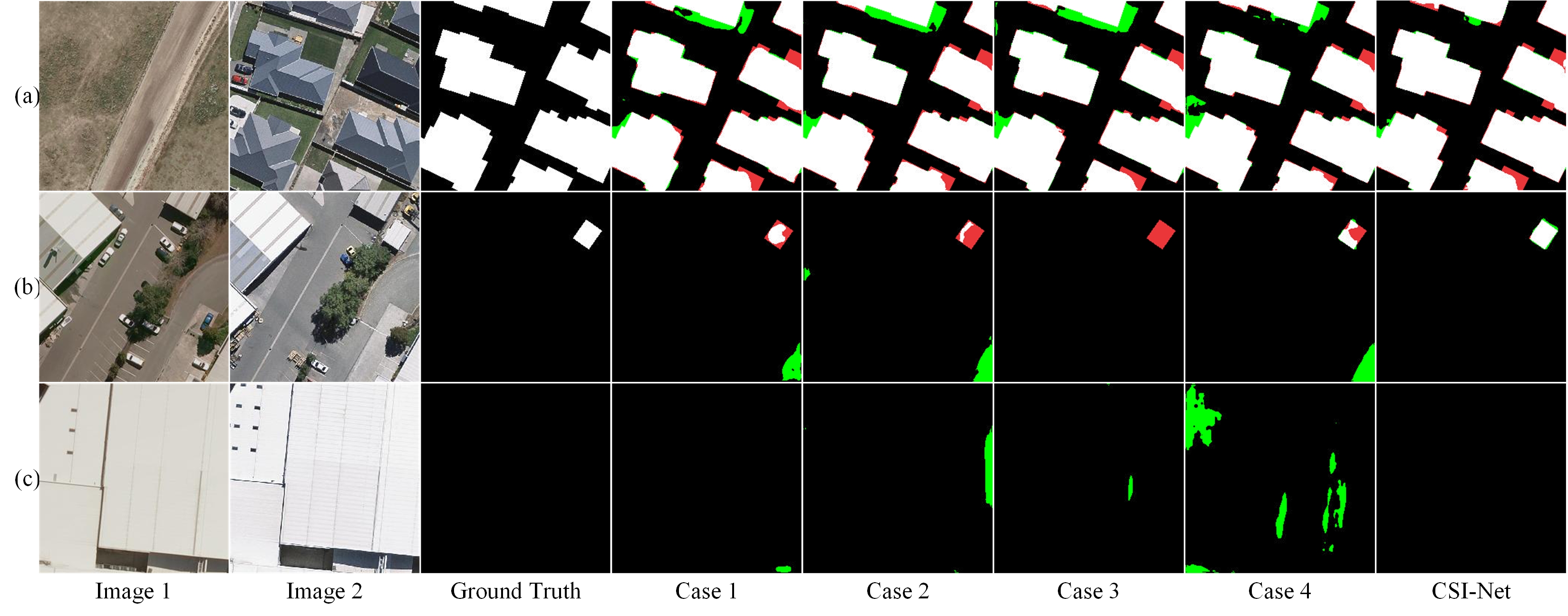}
	\captionsetup{font={footnotesize}}
	\caption{Examples of qualitative results of the ablation study on the WHU-CD dataset.}
	\label{fig8}
	\vspace{-0.2cm}
\end{figure*}

\begin{figure*}[ht]
	\centering
	\includegraphics[scale=0.315]{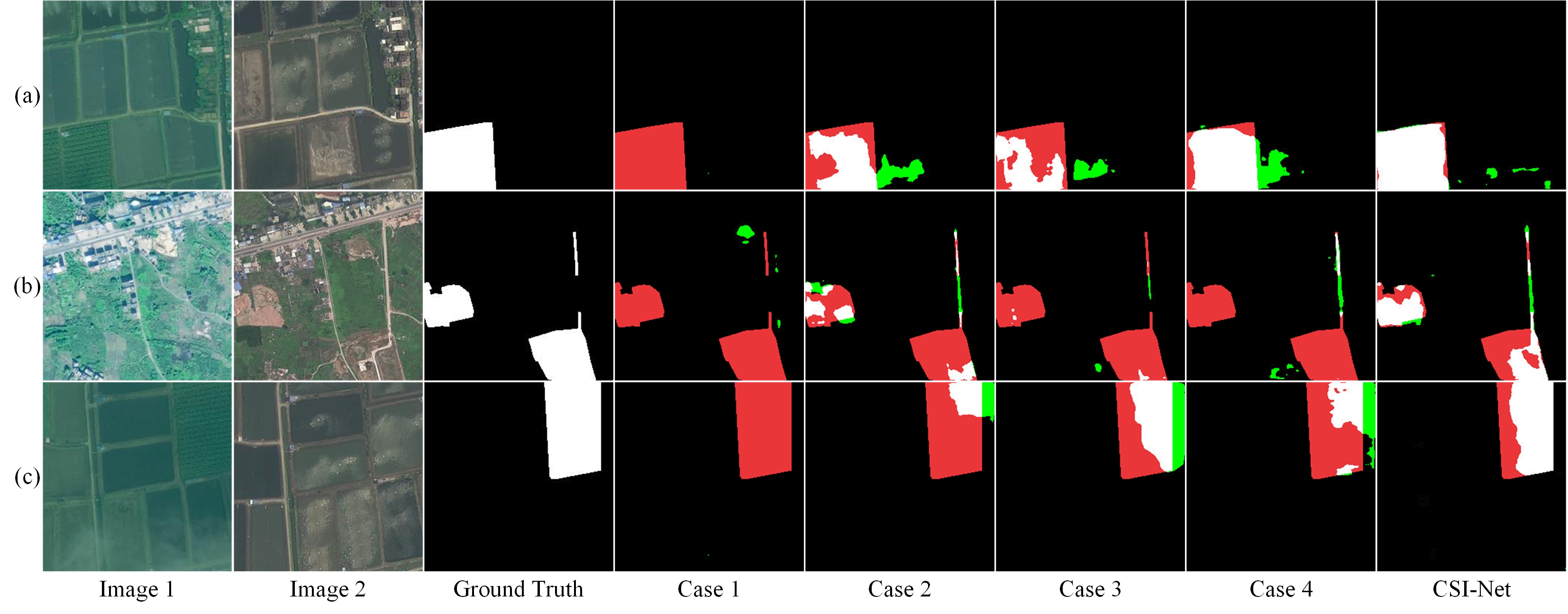}
	\captionsetup{font={footnotesize}}
	\caption{Examples of qualitative results of the ablation study on the CLCD dataset.}
	\label{fig9}
	\vspace{-0.2cm}
\end{figure*}

\begin{figure*}[ht]
	\centering
	\includegraphics[scale=0.22]{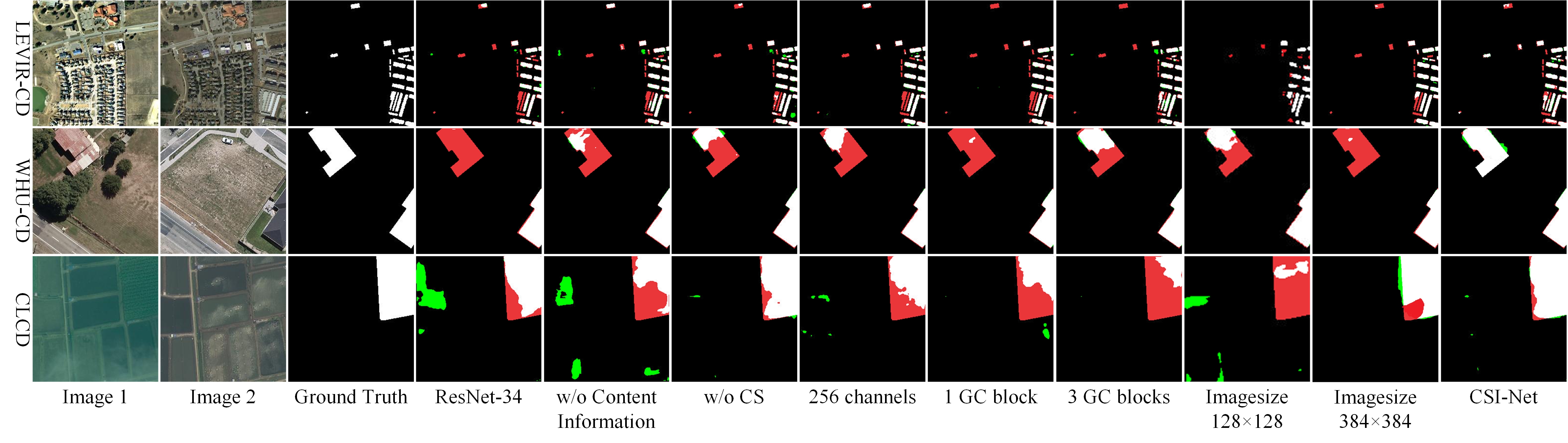}
	\captionsetup{font={footnotesize}}
	\caption{\textcolor{black}{Examples of qualitative results on the analysis of the influences of different network configurations on the three datasets.}}
	\label{fig10}
	\vspace{-0.2cm}
\end{figure*}

\begin{table}[h]
	\centering
	\captionsetup{font={footnotesize}}
	\caption{\textsc{\textcolor{black}{Quantitative Results of All Methods on the CLCD Dataset.}}}
	\footnotesize
	\renewcommand\arraystretch{1.2}
	\begin{tabular}{cccccc}
		\hline
		Metric (\%)  & P              & R              & F1             & IoU            & OA                     \\ \hline
		FC-EF        & 89.04          & 65.03          & 70.90          & 61.54               & 94.38                  \\
		FC-Siam-diff & 85.50          & 73.15          & 77.77          & 67.92               & 94.93                   \\
		FC-Siam-conc & 81.06          & 78.36          & 79.64          & 69.74          & 94.63         \\
		DTCDSCN      & 80.56          & 77.75          & 79.07          & 69.12          & 94.50                  \\
		\textcolor{black}{USSFC-Net}      & \textcolor{black}{85.21}          & \textcolor{black}{79.75}          & \textcolor{black}{82.20}          & \textcolor{black}{72.72}          & \textcolor{black}{95.47}         \\
		BIT          & 80.63          & 75.19          & 77.59          & 67.59          & 94.37                  \\
		SNUNet       & 84.90          & 80.40          & 82.46          & 73.01               & 95.47                  \\
		HANet & 83.30          & 81.47          & 82.36          & 72.85           & 95.20                 \\
		Proposed CSI-Net   & \textbf{85.61} & \textbf{82.23} & \textbf{83.82} & \textbf{74.64} & \textbf{95.75}                  \\ \hline
	\end{tabular}
    \label{tableIII}
\vspace{-0.2cm}
\end{table}

\subsection{Ablation Study}
In this section, to demonstrate the importance of different port of the proposed CSI-Net, we set up the following models and performe ablation experiments on the three considered datasets.

In Case 1, we remove the CGI, SR, and SD modules and only keep ResNet-18 and skip connections. In Case 2, the CGI module is ablated to analyze its influence on all datasets. In Case 3, the SR module in the model of Case 2 is removed to verify the importance of spatial relationships. In contrast to Case 3, we maintain the SR module and eliminate the SD module to define the model of Case 4, in which the contribution of spectral difference is validated.

From the qualitative ablation results in Figs. \ref{fig7}-\ref{fig9}, one can observe that the backbone network is unable to accurately detect the change regions on all three datasets due to the lack of assistance of spatial-spectral information, and a large range of false-negative regions appear. When the CGI module is ablated, more false-positive regions arise in the results because the spatial and spectral information cannot be fused efficiently. The removal of the SR module leads to fragmentation of the change regions and the appearance of irregular false-negative areas. Some buildings similar to the background cannot be accurately detected when SD module is erased. Tables {\ref{tableIV}-\ref{tableVI}} provide the quantitative results obtained in the ablation studies. We can see that the CD results get worse when SR and SD modules are eliminated or replaced, and the proposed CSI-Net achieves the best results in terms of all evaluation metrics.

\begin{table}[h!]
	\centering
	\captionsetup{font={footnotesize}}
	\caption{\textsc{Ablation Study of the Proposed CSI-Net on the LEVIR-CD Dataset.}}
	\scriptsize
	\renewcommand\arraystretch{1.2}
	\begin{tabular}{ccccccccc}
		\hline
		Case & SR & SD &  CGI & P    & R     & F1      & IoU     & OA   \\ \hline
		Case 1 & $\times$      & $\times$ & $\times$             & 85.68   & 66.21   & 71.88   & 62.84   & 96.00 \\
		Case 2 &  $\times$    & \checkmark   & $\times$             & 90.80   & 88.48   & 89.61   & 82.48   & 98.41  \\
		Case 3 & \checkmark    & $\times$ & $\times$               & 93.54 & 87.45  & 90.24 & 83.4  & 98.24     \\
		Case 4 &  \checkmark    & \checkmark & $\times$               & \textbf{94.06} & 89.58  & 91.68 & 85.54  & 98.47     \\
		CSI-Net &  \checkmark    & \checkmark & \checkmark             & 93.98          & \textbf{90.59} & \textbf{92.21} & \textbf{86.34} &  \textbf{98.55} \\ \hline
	\end{tabular}
    \label{tableIV}
	\vspace{-0.2cm}
\end{table}

\begin{table}[h!]
	\centering
	\captionsetup{font={footnotesize}}
	\caption{\textsc{Ablation Study of the Proposed CSI-Net on the WHU-CD Dataset.}}
	\scriptsize
	\renewcommand\arraystretch{1.2}
	\begin{tabular}{ccccccccc}
			\hline
		Case & SR & SD &  CGI & P       & R       & F1      & IoU     & OA   \\ \hline
		Case 1 & $\times$      & $\times$ & $\times$             & 86.38   & 93.85   & 89.72   & 82.66   & 98.09 \\
		Case 2 & $\times$    & \checkmark   & $\times$             & 91.59   & 93.94   & 92.73   & 87.16   & 98.74  \\
		Case 3 & \checkmark    & $\times$ & $\times$               & 93.40 & 95.63  & 94.49 & 89.98  & 99.05     \\
		Case 4 & \checkmark    & \checkmark & $\times$               & 95.48 & 94.25  & 94.86 & 90.60  & 99.15     \\
		CSI-Net & \checkmark    & \checkmark & \checkmark             & \textbf{96.27}          & \textbf{95.43} & \textbf{95.85} & \textbf{92.28} &  \textbf{99.31} \\ \hline
		\end{tabular}
    \label{tableV}
\vspace{-0.2cm}
\end{table}

\begin{table}[h!]
	\centering
	\captionsetup{font={footnotesize}}
	\caption{\textsc{Ablation Study of the Proposed CSI-Net on the CLCD Dataset.}}
	\scriptsize
	\renewcommand\arraystretch{1.2}
	\begin{tabular}{ccccccccc}
			\hline
		Case & SR & SD &  CGI & P       & R       & F1      & IoU     & OA   \\ \hline
		Case 1 & $\times$      & $\times$ & $\times$             & 76.81   & 65.71   & 69.48   & 60.06   & 93.30 \\
		Case 2 & $\times$    & \checkmark   & $\times$             & 77.27   & 79.28   & 78.23   & 68.08   & 93.78  \\
		Case 3 & \checkmark    & $\times$ & $\times$               & 81.37 & 77.63  & 79.35 & 69.45  & 94.65     \\
		Case 4 & \checkmark    & \checkmark & $\times$               & 83.83 & 80.26  & 81.92 & 72.37  & 95.28     \\
		CSI-Net & \checkmark    & \checkmark & \checkmark            & \textbf{85.61} & \textbf{82.23} & \textbf{83.82} & \textbf{74.64} & \textbf{95.75} \\ \hline
		\end{tabular}
    \label{tableVI}
\vspace{-0.2cm}
\end{table}

\begin{table}[]
	\centering
	\captionsetup{font={footnotesize}}
	\caption{\textsc{Comparison Between ResNet-34 and ResNet-18 on the Three Considered Datasets.}}
	\scriptsize
	\renewcommand\arraystretch{1.3}
	\begin{tabular}{ccccccc}
		
		\hline
		
		\multicolumn{1}{l}{Datasets} & ResNet   & P     & R     & F1    & IoU   & OA    \\ \hline
		\multirow{2}{*}{LEVIR}       & ResNet-34 & \textbf{94.21} & 89.42 & 91.66 & 85.50 & 98.47      \\
		& ResNet-18 & 93.98          & \textbf{90.59} & \textbf{92.21} & \textbf{86.34} &  \textbf{98.55} \\ \hline
		
		\multirow{2}{*}{WHU-CD}      & ResNet-34 & 95.27 & 93.08  & 94.14 & 89.42 & 99.04 \\
		& ResNet-18  & \textbf{96.27}          & \textbf{95.43} & \textbf{95.85} & \textbf{92.28} &  \textbf{99.31}  \\ \hline
		
		\multirow{2}{*}{CLCD}        & ResNet-34 & 84.12 &82.14  &83.09  &73.74  &95.48       \\
		& ResNet-18 & \textbf{85.61} & \textbf{82.23} & \textbf{83.82} & \textbf{74.64} & \textbf{95.75} \\ \hline
		
	\end{tabular}
	\label{tableVII}
	\vspace{-0.2cm}
\end{table}


\subsection{Analysis of the Backbone Networks}
In order to explore the effects of different backbone networks on CD results, we replace the backbone ResNet-18 network in encoders with ResNet-34. Table \ref{tableVII} shows the evaluation metrics obtained in the experiments. When ResNet-34 is used as encoders, we can find that the P value increases on both LEVIR-CD and WHU-CD datasets, whereas all other metric values are lower than those of ResNet-18. This is probably because the ResNet-34 has an excessive number of parameters and the backbone networks cannot be trained sufficiently on these datasets. Specifically, ResNet-34 has 21.79M parameters, while ResNet-18 only contains 11.17M parameters. The 4th column of Fig. \ref{fig10} shows the qualitative results of ResNet-34. It can be seen that there are some false-positive and false-negative regions and the overall CD performance decreases.

\subsection{Analysis of the High level Semantic Information}
To facilitate the fusion of spatial and spectral features, we introduce high-level semantic information in the proposed CGI module. To verify the validity of the introduced content information, we performed ablation experiments on the three considered datasets. The evaluation results are shown in Table \ref{tableVIII}, where SI-Net indicates the case in which the content information is not introduced into the CGI module. As shown in the table, SI-Net achieves a better P on the LEVIR-CD dataset, but all other metric values are lower than those obtained by the CSI-Net. This implies that the introduced content information improves the overall CD performance. In both the WHU-CD and CLCD datasets, CSI-Net achieves better results for all metrics except R, which is slightly lower than that of SI-Net. For the WHU-CD dataset, the IoU value is improved from 91.27$\%$ to 92.28$\%$. The introduction of content information raises the value of P by 1$\%$ on the CLCD dataset. The reason for this is that the high-level feature contains rich semantic information, which reduces the gap between spatial and spectral features. As a result, less information is lost during spatial-spectral feature fusion, which effectively improves the CD results.

\subsection{Influence of the Channel Shuffle Operation}
To produce channel-specific attention maps, we introduce the channel shuffle (CS) operation in the CGI module. In this section, the CS operation is ablated to analyze its effectiveness. The results are shown in Table \ref{tableIX}, where w/o CS indicates that CS is removed from the CGI module. As one can see from the table, the CSI-Net achieves the best results on the LEVIR dataset in terms of R, F1, IoU, and OA. In addition, it obtains better values on both the WHU-CD and CLCD datasets. The above results prove that the CS operation helps to improve the spatial-spectral fusion capability of the CGI module. The 6th column of Fig. \ref{fig10} presents an example of qualitative results on the three datasets of CSI-Net without CS. It can be seen that CSI-Net detects more complete CD regions and has lower false positive regions.

\subsection{Analysis of the Number of Channels}
In this  experiments considering the tradeoff between computational complexity and CD performance, we reduce the number of channels of the feature maps in the SD, SR, and CGI modules. As shown in Table \ref{tableX}, when the number of channels is reduced to 256, there is a small increase in P for the LEVIR-CD and WHU-CD datasets, but all other metrics decreased. Besides, the size and FLOPs of model are decreased to 30.993M and 487.83G, respectively. For the CLCD dataset, CSI-Net produces better results in terms of all metrics. The 7th column of Fig. \ref{fig10} illustrates an example of qualitative results on the three datasets for CSI-Net with 256 channels, and the CD results of CSI-Net have fewer false-negative regions. As a result, it can be seen that the proposed CSI-Net is better in terms of overall accuracy.

\subsection{Complexity Analysis}
In this experiment, we compare the computational complexity of different DNNs and illustrate their model size and FLOPs in Table \ref{tableXI}. The results indicate that the proposed CSI-Net has a high FLOPs value, which is mainly caused by the attention calculation in the designed SD and CGI modules. In the SD module, we use GC blocks to fully derive the global information, which also leads to an increase in terms of computational complexity. In the SR and SD modules of CSI-Net, the number of channels is set as 512 to obtain an adequate feature extraction. So, more channels means that the model size of CSI-Net is large.

\begin{figure*}[ht!]
	\centering
	\includegraphics[scale=0.12 ]{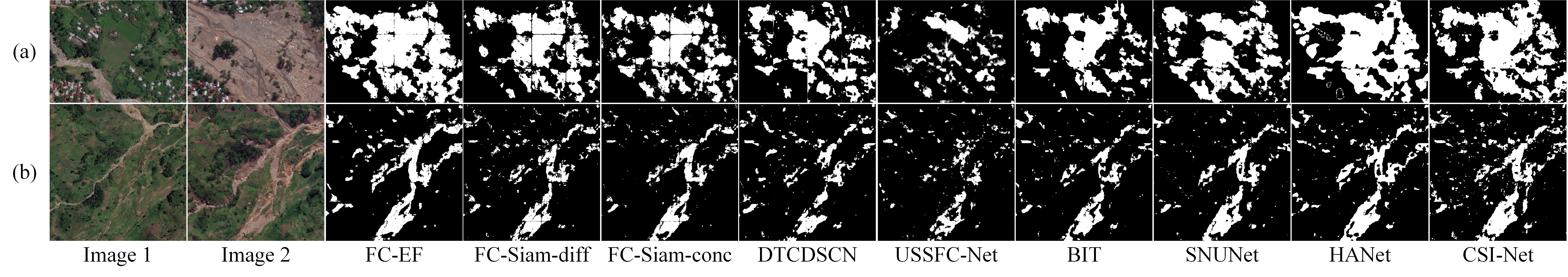}
	\captionsetup{font={footnotesize}}
	\caption{\textcolor{black}{Qualitative comparison of all methods on the real scenarios.}}
	\label{fig11}
	\vspace{-0.2cm}
\end{figure*}

\begin{figure*}[ht!]
	\centering
	\includegraphics[scale=0.336 ]{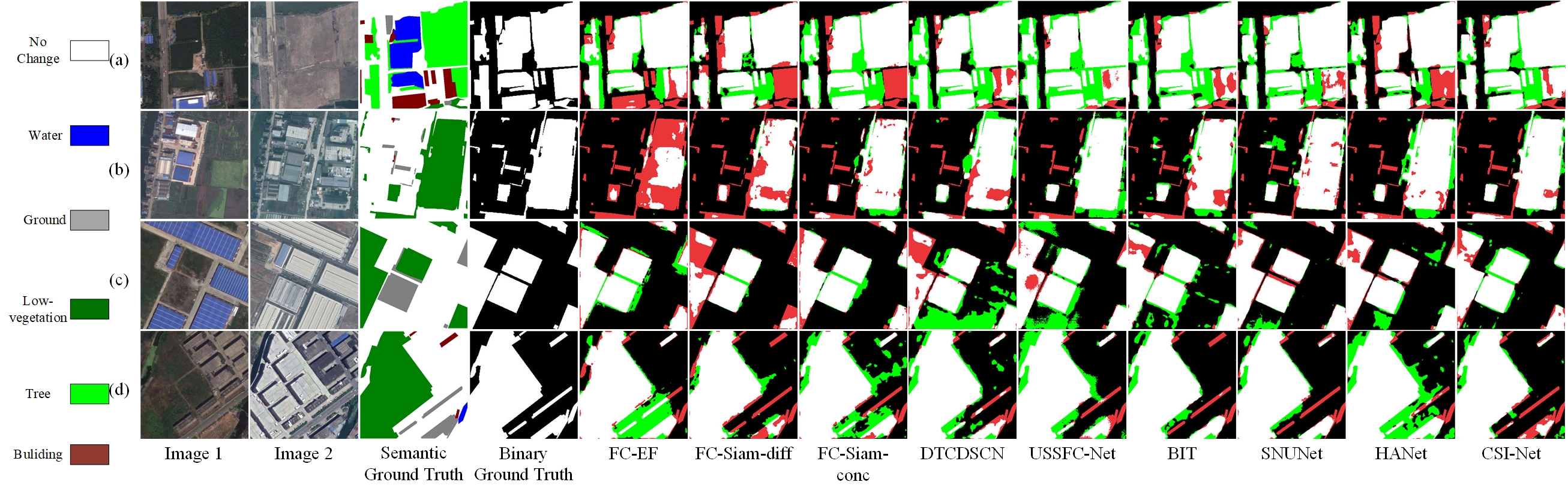}
	\captionsetup{font={footnotesize}}
	\caption{\textcolor{black}{Qualitative comparison of all methods on the Sensetime dataset. (a)-(d): Prediction results of all methods on examples of different image pairs.}}
	\label{fig12}
	\vspace{-0.2cm}
\end{figure*}

\begin{table}[]
	\centering
	\captionsetup{font={footnotesize}}
	\caption{\textsc{Influence of the High Level Semantic Information on the Three Considered Datasets.}}
	\scriptsize
	\renewcommand\arraystretch{1.3}
	\begin{tabular}{ccccccc}
		
		\hline
		
		\multicolumn{1}{l}{Datasets} & Methods   & P     & R     & F1    & IoU   & OA    \\ \hline
		\multirow{2}{*}{LEVIR}       & SI-Net & \textbf{94.12} & 90.39 & 92.16 & 86.27 & 98.55      \\
		& CSI-Net & 93.98          & \textbf{90.59} & \textbf{92.21} & \textbf{86.34} &  \textbf{98.55} \\ \hline
		
		\multirow{2}{*}{WHU-CD}      & SI-Net & \textbf{94.64} & 95.89  & 95.25 & 91.26 & 99.19 \\
		& CSI-Net  & 96.27          & \textbf{95.43} & \textbf{95.85} & \textbf{92.28} &  \textbf{99.31}  \\ \hline
		
		\multirow{2}{*}{CLCD}        & SI-Net & 85.07 & \textbf{82.51}  &83.73  &74.52  &95.68       \\
		& CSI-Net & \textbf{85.61} & 82.23 & \textbf{83.82} & \textbf{74.64} & \textbf{95.75} \\ \hline
		
	\end{tabular}
	\label{tableVIII}
	\vspace{-0.2cm}
\end{table}

\begin{table}[]
	\centering
	\captionsetup{font={footnotesize}}
	\caption{\textsc{Influence of Channel Shuffle on the Three Considered Datasets.}}
	\scriptsize
	\renewcommand\arraystretch{1.3}
	\begin{tabular}{ccccccc}
		
		\hline
		
		\multicolumn{1}{l}{Datasets} & Met`hods   & P     & R     & F1    & IoU   & OA    \\ \hline
		\multirow{2}{*}{LEVIR}       & w/o CS & \textbf{94.33} & 89.5 & 91.76 & 85.63 & 98.49      \\
		& CSI-Net & 93.98          & \textbf{90.59} & \textbf{92.21} & \textbf{86.34} &  \textbf{98.55} \\ \hline
		
		\multirow{2}{*}{WHU-CD}      & w/o CS & 95.87 & 94.28  & 95.06 & 90.94 & 99.18 \\
		& CSI-Net  & \textbf{96.27}          & \textbf{95.43} & \textbf{95.85} & \textbf{92.28} &  \textbf{99.31}  \\ \hline
		
		\multirow{2}{*}{CLCD}        & w/o CS & 85.12 & 81.82  &83.37  &74.10  &95.63       \\
		& CSI-Net & \textbf{85.61} & \textbf{82.23} & \textbf{83.82} & \textbf{74.64} & \textbf{95.75} \\ \hline
		
	\end{tabular}
	\label{tableIX}
	\vspace{-0.2cm}
\end{table}

\begin{table}[]
	\centering
	\captionsetup{font={footnotesize}}
		\caption{\textsc{Influence of the number of channels on the Three Considered Datasets.}}
	\scriptsize
	\renewcommand\arraystretch{1.3}
	\begin{tabular}{ccccccc}
		
		\hline
		
		\multicolumn{1}{l}{Datasets} & Methods   & P     & R     & F1    & IoU   & OA    \\ \hline
		\multirow{2}{*}{LEVIR}       & 256 channels & \textbf{94.26} & 90.19 & 92.12 & 86.19 & 98.54      \\
		& CSI-Net & 93.98          & \textbf{90.59} & \textbf{92.21} & \textbf{86.34} &  \textbf{98.55} \\ \hline
		
		\multirow{2}{*}{WHU-CD}      & 256 channels & \textbf{96.71} & 94.49  & 95.57 & 91.80 & 99.27 \\
		& CSI-Net  & 96.27          & \textbf{95.43} & \textbf{95.85} & \textbf{92.28} &  \textbf{99.31}  \\ \hline
		
		\multirow{2}{*}{CLCD}        & 256 channels & 85.12 & 81.82  &83.37  &74.10  &95.63       \\
		& CSI-Net & \textbf{85.61} & \textbf{82.23} & \textbf{83.82} & \textbf{74.64} & \textbf{95.75} \\ \hline
		
	\end{tabular}
	\label{tableX}
	\vspace{-0.2cm}
\end{table}

\begin{table}[]
	\centering
	\captionsetup{font={footnotesize}}
	\caption{\textsc{\textcolor{black}{Complexity Comparison of Different Methods on the LEVIR-CD Datasets.}}}
	\scriptsize
	\renewcommand\arraystretch{1.4}
	\begin{tabular}{ccccccc}
		\hline
		Methods      & Para.(M)       & FLOPs(G)  \\ \hline
		FC-EF      & 1.351 & 14.308              \\
		FC-Siam-conc    & 1.546   & 21.323        \\
		FC-Siam-diff    & 1.35 & 18.907             \\
		DTCDSCN    & 31.257 & 52.896                  \\
		\textcolor{black}{USSFC-Net}  &\textcolor{black}{2.02} & \textcolor{black}{19.44}                      \\
		BIT    & 3.496 & 42.533                        \\
		SNUNet      & 12.035 & 219.333                  \\
		HANet    & 2.612 & 70.688                        \\
		Proposed CSI-Net      & 57.65 & 695.84                  
		 \\ \hline

	\end{tabular}
	\label{tableXI}
	\vspace{-0.2cm}
\end{table}

\subsection{\textcolor{black}{Experiments on the Other Scenarios}}
\textcolor{black}{In order to verify the effectiveness of the proposed CSI-Net on real scenarios, we conduct some experiments on the CD task containing catastrophic flash floods and mudslides. In these scenarios, the multi-temporal images are captured from the Democratic Republic of the Congo (DRC) on Apr. 10th, 2023 and May 12th, 2023. In addition, the images at time 1 are collected by the WorldView-2 satellite. The images at time 2 are obtained by the WorldView-3 satellite. In this task, there are two pairs of multi-temporal images, whose sizes are 1792×1792 and 1024×768, respectively. Since the image size is too large, we split them into image patches with a size of 256×256 for testing. Then, all CD results of these patches are aggregated to original sizes for demonstration. In this experiment, the DNN models trained on the CLCD dataset are directly used for testing. The results of all methods are shown in Fig.\ref{fig11}. As can be seen in Fig.\ref{fig11}, after the flood, a large area of forests as well as buildings are destroyed, and it can be noticed that all these methods except CSI-Net, SNUNet, and HANet show discontinuities among patches. Besides, it can be seen that CSI-Net can locate the destroyed areas better, which verifies the effectiveness of the proposed method on real scenarios.}

\textcolor{black}{In addition, we select the Sensetime dataset containing 31 “from-to” change types and transform it into a binary CD dataset. The results of all methods are shown in Fig. \ref{fig12}. From Fig. \ref{fig12}, it can be seen that CSI-Net still achieves better results on different change types. Table \ref{tableXIV} provides the results obtained by different methods on this dataset and it can be seen that CSI-Net produces the best results on all the metrics except R. Therefore, the proposed CSI-Net can be applied to the CD task containing different types of surface covers such as forests, land, and buildings.}
\begin{table}[h]\color{black}
	\centering
	\captionsetup{font={footnotesize}}
	\caption{\textsc{\textcolor{black}{Influence of the Size of Images on the Three Considered Datasets.}}}
	\scriptsize
	\renewcommand\arraystretch{1.3}
	\begin{tabular}{ccccccc}
		
		\hline
		
		\multicolumn{1}{l}{Datasets} & Image Sizes   & P     & R     & F1    & IoU   & OA    \\ \hline
		\multirow{3}{*}{LEVIR}       & $128 \times 128$ & 91.73 & 88.03 & 89.78 & 88.74 & 98.11      \\
		&  $256 \times 256$ & 93.98          & \textbf{90.59} & \textbf{92.21} & \textbf{86.34} &  \textbf{98.55} \\
		&  $384 \times 384$ & \textbf{95.19}          & 88.98 & 91.38 & 85.76 &  98.52 \\ \hline
		
		\multirow{3}{*}{WHU-CD}      & $128 \times 128$ & \textbf{97.37} & 94.03 & 95.63 & 91.91 & 99.29      \\
		&  $256 \times 256$ & 96.27          & \textbf{95.43} & \textbf{95.85} & \textbf{92.28} &  \textbf{99.31} \\
		&  $384 \times 384$ & 95.42          & 94.07 & 94.73 & 90.39 &  99.13 \\ \hline
		
		\multirow{3}{*}{CLCD}        & $128 \times 128$ & 84.59 & 79.60 & 81.86 & 72.31 & 95.36      \\
		&  $256 \times 256$ & 85.61          & \textbf{82.23} & \textbf{83.82} & \textbf{74.64} &  \textbf{95.75} \\
		&  $384 \times 384$ & \textbf{85.79}          & 77.40 & 80.92 & 71.28 &  95.35 \\ \hline
		
	\end{tabular}
	\label{tableXII}
	\vspace{-0.2cm}
\end{table}

\begin{table}[]\color{black}
	\centering
	\captionsetup{font={footnotesize}}
	\caption{\textsc{\textcolor{black}{Influence of the Number of GC blocks on the Three Considered Datasets.}}}
	\scriptsize
	\renewcommand\arraystretch{1.3}
	\begin{tabular}{ccccccc}
		
		\hline
		
		Datasets & GC blocks Numbers   & P     & R     & F1    & IoU   & OA    \\ \hline
		\multirow{3}{*}{LEVIR}       & 1 & 93.24 & 84.24 & 88.15 & 80.15 & 97.95      \\
		&  2 & 93.98          & \textbf{90.59} & \textbf{92.21} & \textbf{86.34} &  \textbf{98.55} \\
		&  3 & \textbf{94.51}          & 89.40 & 91.78 & 86.34 &  98.54 \\ \hline
		
		\multirow{3}{*}{WHU-CD}      & 1 & \textbf{96.64} & 93.78 & 95.16 & 91.11 & 99.21      \\
		&  2 & 96.27          & \textbf{95.43} & \textbf{95.85} & \textbf{92.28} &  \textbf{99.31} \\
		&  3 & 95.75          & 94.69 & 95.21 & 91.19 &  99.22 \\ \hline
		
		\multirow{3}{*}{CLCD}        & 1 & 84.30 & 82.22 & 83.22 & 73.89 & 95.52      \\
		&  2 & \textbf{85.61}          & \textbf{82.23} & \textbf{83.82} & \textbf{74.64} &  \textbf{95.75} \\
		&  3 & 84.73          & 78.42 & 81.19 & 71.56 &  95.29 \\ \hline
		
	\end{tabular}
	\label{tableXIII}
	\vspace{-0.2cm}
\end{table}

\begin{table}[h]\color{black}
	\centering
	\captionsetup{font={footnotesize}}
	\caption{\textsc{\textcolor{black}{Quantitative Results of All Methods on the Sensetime Dataset.}}}
	\footnotesize
	\renewcommand\arraystretch{1.2}
	\begin{tabular}{cccccc}
		\hline
		Metric (\%)  & P              & R              & F1             & IoU            & OA                     \\ \hline
		FC-EF        & 77.42          & 65.60          & 68.03          & 55.21               & 80.88                  \\
		FC-Siam-diff & 83.28          & 64.27          & 66.79          & 54.41              & 81.56                   \\
		FC-Siam-conc & 78.95         & 68.45          & 71.14          & 58.23          & 82.14         \\
		DTCDSCN      & 79.10         & 75.42          & 76.93          & 64.26          & 83.93                  \\
		USSFC-Net      & 74.93          & 75.95          & 75.41          & 62.25          & 82.11         \\
		BIT          & 79.00          & 74.77          & 76.46          & 63.73          & 83.75                  \\
		SNUNet       & 77.87          & 74.72          & 76.04          & 63.16               & 83.19                  \\
		HANet & 69.24          & \textbf{76.54}          & 71.48          & 58.37           & 81.54                 \\
		Proposed CSI-Net   & \textbf{79.87} & 75.53 & \textbf{77.27} & \textbf{64.71} & \textbf{84.31}                  \\ \hline
	\end{tabular}
	\label{tableXIV}
	\vspace{-0.2cm}
\end{table}

\begin{table}[h]\color{black}
	\centering
	\captionsetup{font={footnotesize}}
	\caption{\textsc{\textcolor{black}{Quantitative Results of the Generalization Ability of All Methods on the WHU-CD and LEVIR-CD Datasets.}}}
	\footnotesize
	\renewcommand\arraystretch{1.2}
	\begin{tabular}{cccccc}
		\hline
		Dataset       &\multicolumn{2}{c}{LEVIR-CD}  &\multicolumn{2}{c}{WHU-CD}        \\ \hline
		Metric (\%)  & F1              & IoU              & F1             & IoU                                 \\ 
		FC-EF        & 49.74          & 48.00          & 66.95          & 57.06                                \\
		FC-Siam-diff & 49.96          & 48.09          & 71.19          & 61.31                                 \\
		FC-Siam-conc & 48.63        & 47.44          & 70.9          & 61.9                 \\
		DTCDSCN      & 49.31         & 47.77          & 69.72          & 60.55                            \\
		USSFC-Net      & 49.05          & 47.18          & 59.19          & 50.87                  \\
		BIT          & 49.12          & 47.61          & 61.30         & 52.98                           \\
		SNUNet       & 50.05          & 47.87         & 68.38          & 59.64                                \\
		HANet & 49.11          & 47.11          & 67.16          & 58.43                            \\
		Proposed CSI-Net   & \textbf{50.24} & \textbf{48.16} & \textbf{75.28} & \textbf{65.57}                  \\ \hline
	\end{tabular}
	\label{tableXV}
	\vspace{-0.2cm}
\end{table}
\begin{figure*}[ht!]
	\centering
	\includegraphics[scale=0.3 ]{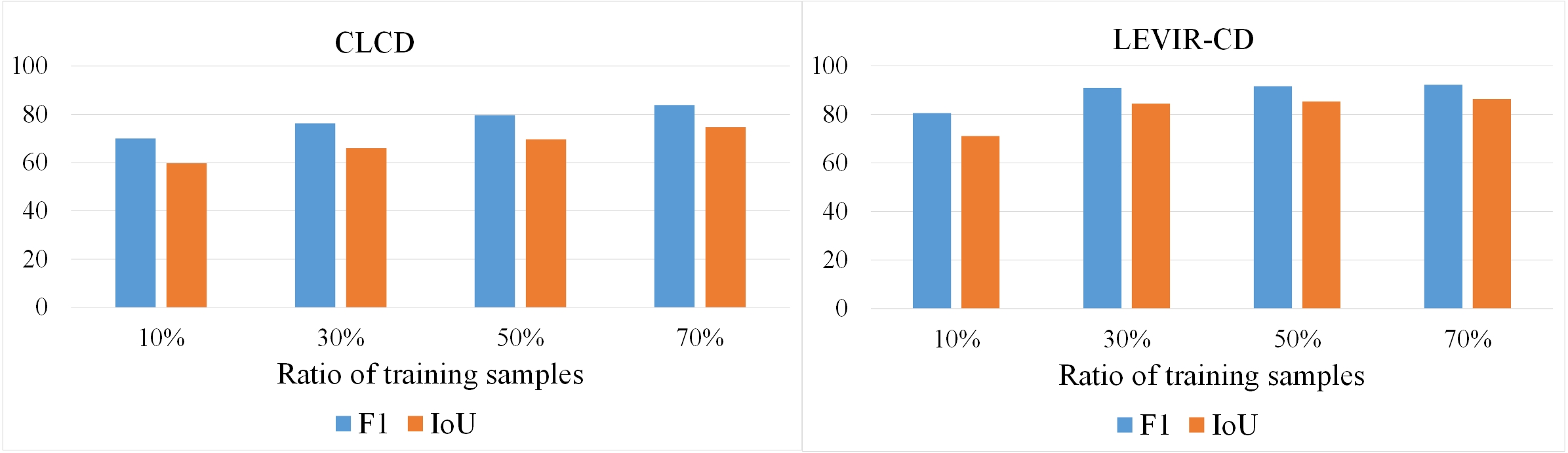}
	\captionsetup{font={footnotesize}}
	\caption{\textcolor{black}{F1$\%$ and IoU$\%$ results based different proportions of the training datasets, such as 70$\%$, 50$\%$, 30$\%$, and 10$\%$}}
	\label{fig13}
	\vspace{-0.2cm}
\end{figure*}

\begin{figure*}[ht!]
	\centering
	\includegraphics[scale=0.35 ]{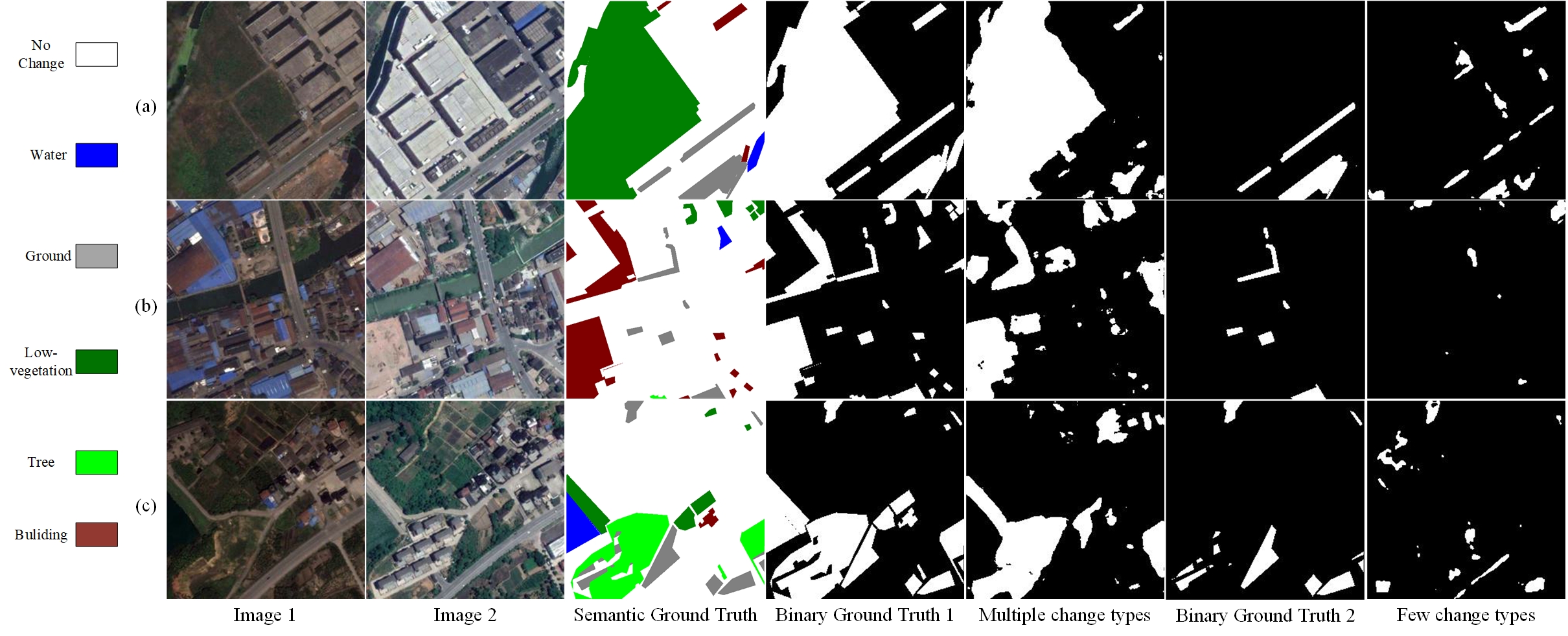}
	\captionsetup{font={footnotesize}}
	\caption{\textcolor{black}{Example of the analysis for the different training sample types }}
	\label{fig14}
	\vspace{-0.2cm}
\end{figure*}
\subsection{\textcolor{black}{Analysis of Parameters}}
	\textcolor{black}{In order to explore the effect of different parameters on CD accuracy, we conduct extensive experiments on the size of training samples and the number of GC blocks in the SR module. The experimental results are shown in Table \ref{tableXII} and Table \ref{tableXIII}. As can be seen from Table \ref{tableXII}, training samples with larger sizes do not bring better CD results . When the size of the image increases, some disturbing information will also become more obvious, which have an impact on the CD results. When the image size is reduced, more spatial and spectral details are lost, resulting in many smaller targets not being detected. In Fig. \ref{fig10}, the 10th and 11th columns show the results of CSI-Net trained on samples with different sizes. It can be seen that after either increasing or decreasing the image size, more undetected regions appear in the results.}
	
	\textcolor{black}{Table \ref{tableXIII} shows the results achieved by CSI-Net with different GC Block numbers on the three datasets. It can be seen that when the number of GC blocks increases or decreases, the obtained CD results become worse. When GC Block=1, the SR module is unable to comprehensively model the overall spatial relationships, so the obtained results will be worse. When the number of GC blocks = 3, the CD results also deteriorate, which may be due to the disappearance of the gradient caused by the excessive stacking of graph convolution layers. The 8th and 9th columns in Fig. \ref{fig10} show the results by changing the number of GC blocks. It can be seen that there are more red areas in the results of CSI-Net with 1 or 3 GC blocks.}
	
\subsection{\textcolor{black}{Analysis of Training Samples}}
\textcolor{black}{In this section, we conduct experiments to analyze the influences of change types and the number of training samples on the proposed CSI-Net. First, the number of training samples decreases to 50$\%$, 30$\%$, and 10$\%$ from 70$\%$. Fig. \ref{fig13} illustrates the F1 and IoU values of CSI-Net on CLCD and LEVIR-CD datasets. As shown in Fig. \ref{fig13}, when the number of training samples decreases, both F1 and IoU of CD decrease. Meanwhile, the values of F1 and IoU of 30$\%$ and 50$\%$ on the LEVIR-CD datasets also prove that the CD performance of the proposed CSI-Net is more stable.}

\textcolor{black}{Besides, Fig. \ref{fig14} demonstrates the experimental results in terms of change types. In this experiment, all 31 kinds of change are first used for training and the CD results of the model of CSI-Net trained on all change types are shown in the fifth column of Fig. \ref{fig14}. In addition, CSI-Net is also trained on 2 change types and the corresponding test results are given in the seventh column of Fig. \ref{fig14}. From Fig. \ref{fig14}, we can see the change types of training samples determines the types that the trained model can detect. when we remove the type of building changes, the proposed CSI-Net will not detect building changes anymore. Thus, the proposed CSI-Net can detect the interested change types, and ignore the influence of other irrelevant changes on the CD task.}
\subsection{\textcolor{black}{Analysis of Generalization Ability}}
\textcolor{black}{To explore the generalization ability of the proposed CSI-Net, we first train CSI-Net on the WHU-CD dataset and then the obtained model is tested on the LEVIR-CD dataset. Similarly, the CSI-Net trained on the LEVIR-CD dataset is also tested on the WHU-CD dataset. Table \ref{tableXV} lists the results of all methods in terms of generalization ability. From this table, we can see that all methods suffer from performance degradation due to the distribution differences between the two datasets. However, it can be observed that the proposed CSI-Net still obtains the best values compared to other methods, which demonstrates the better generalization ability of CSI-Net.}

\section{Conclusion}
In this paper, we have proposed a CSI-Net for the CD of multi-temporal images. In CSI-Net, global image information is modeled by the SR module to efficiently locate the change regions. Meanwhile, the SD module exploits the spectral differences between different images to mitigate the influence of pseudo-change areas. In addition, in order to avoid information loss during feature fusion, we also design a CGI module to guide the aggregation among spatial-spectral features and content information. The effectiveness of the proposed CSI-Net is verified on LEVIR-CD, WHU-CD, and CLCD datasets. \textcolor{black}{This indicates that the proposed CSI-Net is suitable for many types of change detection.} Compared to eight state-of-the-art DNNs, CSI-Net produces better CD accuracy. Despite the satisfactory performance, the proposed CSI-Net is limited by the large model and high FLOPs. For future work, we will explore more effective modules to capture the spatial relationships and spectral differences in multi-temporal images.

\bibliographystyle{IEEEtran}
\bibliography{Reference}

\end{document}